\documentclass{article}
\usepackage[dblblindworkshop,final]{neurips_2026}
\workshoptitle{VLM4RWD}

\usepackage[utf8]{inputenc}
\usepackage[T1]{fontenc}
\usepackage{hyperref}
\usepackage{url}
\usepackage{booktabs}
\usepackage{amsfonts}
\usepackage{amssymb}     
\usepackage{nicefrac}
\usepackage{microtype}
\usepackage{graphicx}
\usepackage{amsmath}
\usepackage{multirow}
\usepackage{siunitx}
\usepackage{xcolor}

\newcommand{\ctext}{\ensuremath{c_{\mathrm{text}}}}
\newcommand{\cbbox}{\ensuremath{c_{\mathrm{bbox}}}}

\title{VERDICT: Visual Extraction-Result Decoder with Integrated Confidence
and Traceable Evidence}

\author{%
  Harikrishnan P M \quad Goutham Vignesh \quad Siva Prasad \quad Zishan Ahmad \\[0.5ex]
  \textbf{Mahesh Basveshwar Deshmukhe} \quad \textbf{Varun V} \quad \textbf{Rohit Agrawal} \quad \textbf{Vishal Vaddina} \\[1.5ex]
  Phi Labs, Quantiphi \\
  \texttt{harikrishnan.m@quantiphi.com}
}

\begin{document}

\maketitle
\begin{abstract}
Enterprises run schema-guided structured field extraction on documents with frontier multimodal
models, and accuracy keeps improving as new models and systems arrive. Two things have not kept pace: the
evidence for each extracted value, meaning the location of the document it was read from, and how far
to trust the value and that evidence. Neither is a component of most extraction systems today, so we build them as a framework that attaches to an existing extractor and leaves it untouched. We present \textbf{VERDICT}, a plug-in framework that uses a small Vision--Language Model (VLM) to ground/localize the evidence once per document over a fixed external extractor's output, then calibrates value and evidence confidence. Fine-tuning the VLM for evidence localization substantially improves visual evidence quality and, more importantly, expands VERDICT's selective-prediction operating range. At a strict confidence threshold $0.80$,
VERDICT with the fine-tuned VLM gives usable precision--coverage trade-offs for both value and
evidence confidence across the in-distribution and out of distribution datasets, whereas the frozen setting often collapses to little or no coverage. Running an
open-weight VLM for localization rather than a closed API makes its internal states available to the calibrator,
and they are worth more than output log-probabilities: in-distribution AUROC rises from $0.730$ to
$0.914$ for value confidence and from $0.681$ to $0.906$ for evidence confidence. This advantage does
not survive a distribution shift. We show that the signal is displaced rather than destroyed, and that
labeling roughly $120$ fields from the target dataset takes evidence confidence from $0.641$ to
$0.905$ on one dataset and from $0.675$ to $0.829$ on the other.
\end{abstract}

\section{Introduction}
\label{sec:intro}

Schema-guided structured field extraction from business documents is in production, and its
accuracy is under active competition from both frontier multimodal models and the extraction pipelines built around
them \citep{idpleaderboard,barzelay2026varex,ji2026unikie,zhang2026extractbench,longextractionbench}. Two things have not moved with it, and together they are what make an extracted value usable
without a person reading the whole page: its evidence, by which we mean the location in the document the value was
read from, and a measure of how far to trust the value and its evidence. Document extraction is therefore a real-world Vision Language Model (VLM) reliability setting where correctness is insufficient without grounded visual evidence and calibrated risk. A field accepted without review has to be checkable
afterwards, and an auditor cannot check a value without an evidence attached
\citep{perot2024lmdx,mortezaagha2025provenance,ahuja2026evisearch}. A field
sent for review costs money, so a deployment should send only those that need it, which requires a
score a threshold can be set on \citep{geifman2017selective,srinivasan2024selective}.

Localizing the evidence for an extracted value has been studied as a task of its own. One work frames it as generation, emitting coordinates as text so that a language model returns each value together with its position \citep{perot2024lmdx}. Another optimises a VLM to answer a question about a document and mark the region supporting that answer
\citep{pm2026perception}. Both produce evidence. Neither statement says whether the evidence it
produced is right. Confidence for document extraction has been studied separately and separately from evidence.
Systems fuse output token probabilities with document-side signals, or return prediction sets
that carry coverage guarantees \citep{kumar2026extractconf,macdonald2025embedding}.
All of them score the extracted value. None of them score the evidence, as none produces evidence that is its own to score.

Keeping the two apart matters, because they fail apart: a value can be correct while the evidence
located for it is wrong and the reverse, so one number cannot tell a practitioner which exposure
they have. A deployment therefore needs
three things at once: \textbf{(i)} evidence for every field of a schema, obtained in one pass
rather than one query per field; \textbf{(ii)} confidence that is calibrated, so a threshold means
what it says; and \textbf{(iii)} that confidence separated into the value and its evidence, since
those are the two ways the field can fail. No published system provides all three.

\begin{figure}[t]
\centering
\includegraphics[width=\linewidth]{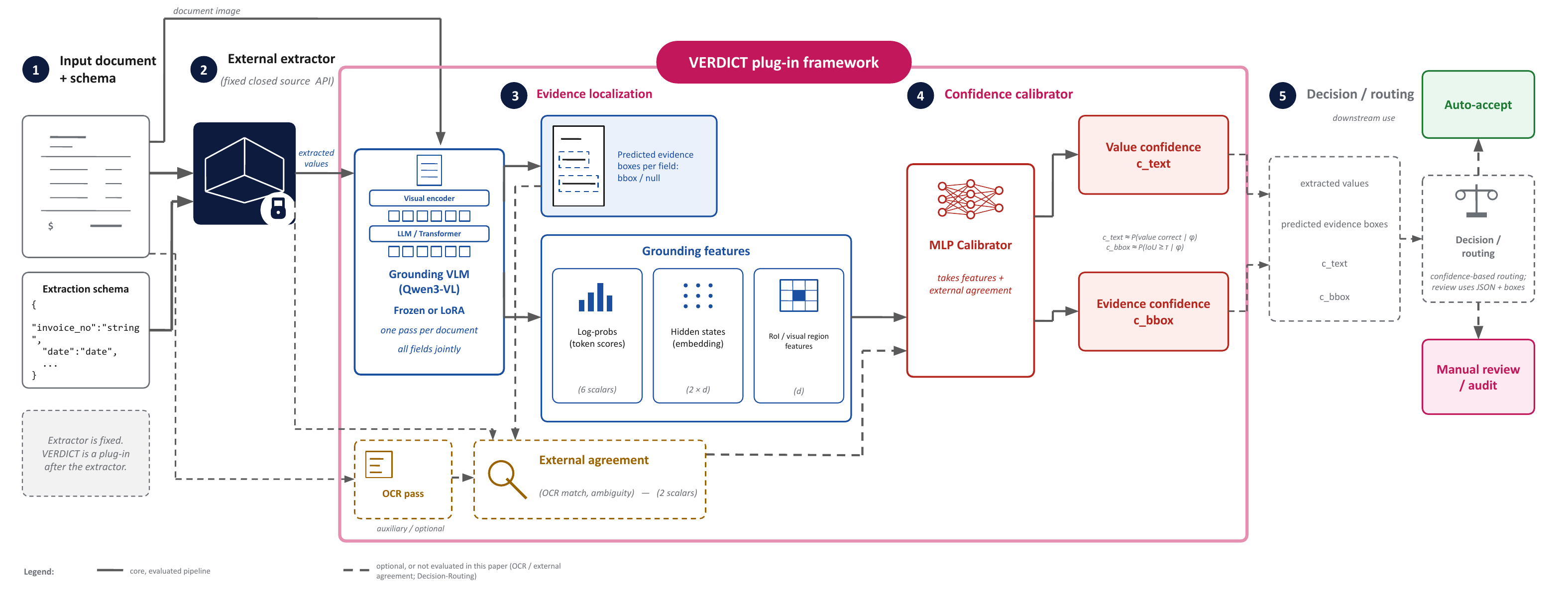}
\caption{The VERDICT framework: A fixed external extractor returns a value per field; a
VLM localizes the evidence for each value in one pass. That single pass yields
both the location, which is shipped to the reviewer as evidence, and the features from which two
confidences are calibrated. Feature counts are for the full configuration (zoom in for best view).}
\label{fig:arch}
\end{figure}

Most extraction pipelines in use today do not carry grounding/localization or confidence as components, and
adding either means changing the part of the stack that was chosen for extraction quality alone. Therefore, we attach both to an existing extractor and leave it untouched. We present \textbf{VERDICT} as shown in Figure~\ref{fig:arch}, a plug-in framework that uses a small VLM to localize evidence once per document over a fixed external extractor's output, then calibrates value and evidence confidence. For every requested field, the framework localises the evidence region on which a reviewer or auditor can act on, and estimates two calibrated probabilities: that the value is correct and that the evidence box is correct. Because the VLM is open-weight, its internal states are
available to the calibrator, which is not the case when confidence has to be built from whatever an extraction API exposes.

\paragraph{Contributions.}
\begin{enumerate}
\itemsep1pt
\item A plug-in localization and calibration framework that attaches to a fixed external extractor,
  localises every field of a schema in a single forward pass, and estimates separate value and
  evidence confidence from VLM features plus optional external agreement.
\item Internal states probed for confidence in a model that emits evidence rather than an answer,
  with a second probability calibrated about whether that evidence is right. They are worth $0.184$
  and $0.225$ AUROC over output log-probabilities on the two targets in distribution, and are
  readable only because the VLM is open-weight.
\item The features that calibrate in distribution are not those that calibrate under distribution
  shift. We give the mechanism: the signal is displaced, not destroyed.
\item What it costs to recover calibration on a new dataset, roughly $120$ labelled fields; why a
  scalar metric can report our least usable setting as our best, so precision is never read without
  its coverage; and a deployment decision table derived from the results rather than asserted.
\end{enumerate}

\section{Related work}
\label{sec:related}

\paragraph{Extraction and evidence localization.}
Returning an extracted value together with the region it was read from has been posed in two ways
that differ mainly in granularity. One asks the model a single question at a time and requires it to
mark the region supporting its answer, placing the task close to visual question answering
\citep{venna2026m3grounder}; these systems are trained by supervised fine-tuning on large annotated
datasets built for the purpose \citep{zhou2025dogr,venna2026m3grounder}, or by reinforcement
learning against overlap-based rewards \citep{pm2026perception}. The other is schema-guided: the
model receives the full set of fields to populate and returns values and positions for all of them
in one pass, with layout communicated to a language model by encoding coordinates as text and
completions mapped back to document segments \citep{perot2024lmdx}. In both, the location is produced and then scored against ground truth afterwards. None
attaches a forward-looking statement of how far the value or that location can be trusted.

\paragraph{Confidence calibration.}
These methods differ in what they read. The most direct asks the model to state its confidence in words
\citep{yoon2025reasoning}, which can also be trained explicitly \citep{li2025conftuner}. The most common reads the output distribution, through
token log-probabilities or agreement across samples: the default in deployed systems, and found
insufficient, since sequence probability tracks length and word frequency as much as correctness
\citep{azaria2023internal} and confidently wrong outputs still score highly
\citep{macdonald2025embedding}. A third line probes what the model represents rather than what it emits: classifiers on hidden
activations separate true from false statements better than the model's own sentence probability
\citep{azaria2023internal,beigi2024internalinspector,yang2026trust}. VLMs add a specific failure: a fluent, confident answer can come from language
priors with the image contributing nothing, which a probe detects by penalising confidence that
survives blanking the image \citep{khanmohammadi2026bicr}. Internal representations also attribute
errors to perception or knowledge before generation \citep{vo2026attributing}, and calibration has
been studied in high-stakes multimodal settings \citep{du2026medvqa}. Within document extraction, one system fits a classifier to the hidden states of an open-weight
extractor prompted to critique its own output \citep{macdonald2025embedding}. Another fuses output statistics with OCR agreement, image quality
and layout \citep{kumar2026extractconf}. Both score the value alone; neither produces
evidence, so neither can say whether the region a value came from is right, and where the extractor
is reached through an API its internal states cannot be read.

\section{Method}
\label{sec:method}

\subsection{Task formulation}
\label{sec:method-task}

A document is an image $D$ with a schema $F = \{f_1, \dots, f_n\}$ of field names to populate. A
fixed external extractor $E$, treated as a black box and never modified, returns a value for each
field or null where it judges the field absent.
\begin{equation}
E(D, F) \rightarrow \{\hat{v}_f\},
\qquad
G\big(D, \{(f, \hat{v}_f)\}\big) \rightarrow \{\hat{b}_f\},
\qquad
C(\phi_f) \rightarrow \big(c_{\mathrm{text},f},\; c_{\mathrm{bbox},f}\big)
\end{equation}
over $f \in F$, where $G$ is a VLM that localizes evidence, referred to throughout as the grounder,$\hat{b}_f = (\hat{x}_1, \hat{y}_1, \hat{x}_2, \hat{y}_2) \in [0,1000]^4 \cup \{\varnothing\}$ is the
evidence location for field $f$ or $\varnothing$ when it cannot be located, and $\phi_f$ are features
read from the same forward pass of $G$ that produced $\hat{b}_f$.

The two probabilities target two different failures:
\begin{equation}
y^{\mathrm{text}}_f = \mathbb{1}\!\left[\mathrm{norm}(\hat{v}_f) = \mathrm{norm}(v_f)\right],
\qquad
y^{\mathrm{bbox}}_f = \mathbb{1}\!\left[\mathrm{IoU}(\hat{b}_f, b_f) \geq \tau\right]
\end{equation}
with $\mathrm{norm}$ lower-casing and stripping whitespace, $\tau$, and two calibrated
targets:
\[
c_{\mathrm{text},f} \approx P(y^{\mathrm{text}}_f = 1 \mid \phi_f),
\qquad
c_{\mathrm{bbox},f} \approx P(y^{\mathrm{bbox}}_f = 1 \mid \phi_f).
\]

One asymmetry governs how the tables are read. $y^{\mathrm{text}}$ is fixed by the extractor, which
we hold constant, so it does not change when the grounding model changes. $y^{\mathrm{bbox}}$ is
defined by the grounder under test, so the same field can be labelled positive under one grounder
and negative under another, and evidence-confidence figures are compared down a column rather than
across grounders.

\subsection{The localization pass and its features}
\label{sec:method-pass}

The extracted values are presented to $G$ as a JSON object keyed by field name,
$\{f : \hat{v}_f\}_{f \in F}$, and it returns an object of the same shape,
$\{f : \hat{b}_f\}_{f \in F}$, with null where a field cannot be located. Putting the whole schema
in one prompt keeps the cost at a single pass regardless of schema size, and gives each field's
coordinates their own token positions inside one generated sequence. We use one backbone in two
conditions, frozen weights and the same model fine-tuned for evidence localization, and every result
below is reported for both.

Four feature groups are captured while $G$ runs, differing in what they can see. Each field gets its
own copy of all four, read from the tokens where the model wrote that field's coordinates, so a
schema of $n$ fields gives $n$ separate feature vectors out of the one pass.
\textbf{Output log-probabilities (6 scalars):} mean, minimum, standard deviation and count over that
field's coordinate tokens, plus flags for extractor abstention and an empty region. This is what a
closed API would expose. \textbf{Hidden states ($2 \times d$):} what the model was computing inside
itself as it wrote those coordinates, averaged across them, read at two layers, the last layer where
visual features are injected and one near the end (Appendix~\ref{app:repro}).
\textbf{Visual region features ($d$):} visual encoder features pooled over the predicted region,
which asks what is actually present where the model pointed rather than what it said.
\textbf{External agreement (2 scalars):} a predicted region is checkable in a way a generated
string is not, so we run OCR, take the words inside $\hat{b}_f$, and compute a fuzzy match against
$\hat{v}_f$ together with a count of how many distinct places in the document contain that value.
Existing confidence work uses OCR to \emph{construct} a region \citep{kumar2026extractconf}; here OCR
is only a post-localization check: the region already exists and OCR \emph{tests} it, which is only possible because the model produces a
pointer of its own.

\subsection{Calibrator and training}
\label{sec:method-calibrator}

$C$ is a small multi-layer perceptron with a shared trunk and two independent linear heads, in two
sizes chosen by which feature groups a configuration uses: MLP\textsubscript{A} for the scalar groups
alone ($6 \to 64 \to 32$), and MLP\textsubscript{B} once a high-dimensional group enters, projecting
each such group to $256$ dimensions, concatenating them with the scalars and passing the result
through $\cdot \to 512 \to 256 \to 128$, with LayerNorm, GELU and dropout throughout. Training
minimises binary cross-entropy summed over the two heads with equal weight, over fields that have
ground truth; class-frequency weighting was tested and rejected, giving no ranking gain and much
worse calibration. Four configurations instantiate VERDICT throughout \S\ref{sec:results}. \textbf{M1} is a trained
calibrator over log-probability statistics and null flags, distinct from the untrained output-probability baseline. \textbf{M2} adds the two external agreement features, OCR match score and ambiguity count, so
$\text{M2}-\text{M1}$ isolates external evidence. \textbf{M3} adds hidden states and region
features, so $\text{M3}-\text{M1}$ isolates model internals. \textbf{M4} combines all feature groups and is the all-feature instantiation.

\section{Experimental setup}
\label{sec:setup}

\subsection{Data}
\label{sec:setup-data}

In-distribution (ID) training spans both KIE categories the benchmark literature distinguishes
\citep{yang2024ccocr,ji2026unikie}: \emph{constrained-category}, where a schema predefines the keys,
supplied by \emph{DocILE} \citep{simsa2023docile} under a fixed 33-field schema identical across
documents; and \emph{open-category}, where the keys are whatever is present on the page, supplied by
\emph{KVP10k} \citep{naparstek2024kvp10k}, whose field names are page text and so differ from one
document to the next. Training across both denies the grounder and the calibrator any field name they can memorise.

Out-of-distribution (OOD) evaluation uses two unrelated form datasets,
\emph{Form-NLU} \citep{ding2023formnlu} (substantial-holder notices, 12 fields) and \emph{VRDU registration-form} \citep{wang2023vrdu} (foreign-agent filings, 5 fields). Table~\ref{tab:data} gives every split. The grounder's fine-tuning documents, the calibrator's
training documents and the evaluation sets are mutually disjoint. A target's \emph{base rate} is the fraction of annotated fields whose label is $1$; the two base
rates are what the calibrator's heads are fitted against.

\begin{table}[t]
\centering\small
\caption{An annotated field is one field of one document for which ground truth exists.
$y_{\mathrm{text}}$ is fixed by the extractor and does not vary with the grounder;
$y_{\mathrm{bbox}}$ is defined by the grounder under test and does vary. The fine-tuning row is trained
on regions, so it has no field-level columns.}
\label{tab:data}
\setlength{\tabcolsep}{5pt}
\begin{tabular}{llrrrrr}
\toprule
& & & Annotated & & \multicolumn{2}{c}{$y_{\mathrm{bbox}}$} \\
\cmidrule(l){6-7}
Data & Purpose & Documents & fields & $y_{\mathrm{text}}$ & frozen & LoRA \\
\midrule
DocILE + KVP10k & grounder fine-tuning   & 10{,}500 & ---        & ---   & ---   & ---   \\
DocILE + KVP10k & calibrator training    &  8{,}840 & 75{,}469   & 0.593 & 0.432 & 0.848 \\
DocILE + KVP10k & ID evaluation          &  1{,}883 & 15{,}291   & 0.600 & 0.421 & 0.852 \\
FormNLU         & OOD evaluation         &     535  &  6{,}229   & 0.772 & 0.019 & 0.586 \\
VRDU            & OOD evaluation         &     200  &     501    & 0.864 & 0.421 & 0.699 \\
\bottomrule
\end{tabular}
\end{table}


\subsection{Models and training}
\label{sec:setup-models}

The extractor is Gemini 3 Flash (API identifier \texttt{gemini-3-flash-preview})\ \citep{geminiteam2023gemini}, used through its API and never modified. The grounder is
Qwen3-VL-4B-Instruct \citep{qwen3vl} in two conditions: frozen weights, and LoRA \citep{hu2021lora}
fine-tuned for evidence localization with rank $16$, $\alpha = 16$, adapters on all attention and MLP
projections, and one epoch. Fine-tuning optimises localization only; the extracted values are inputs,
never targets.

The calibrator uses Adam, batch size $64$, at most $80$ epochs with early stopping, and learning rate
$3{\times}10^{-4}$ for MLP\textsubscript{A} and $1{\times}10^{-4}$ for MLP\textsubscript{B}. Every
configuration is trained with five seeds, and a difference is reported as a finding only when it
exceeds the seed-to-seed spread.

\subsection{Baselines and metrics}
\label{sec:setup-baselines}

Two baselines. \textbf{OCR mapping} is the alternative to a grounder: run OCR, find the
extracted value's occurrences on the page, and resolve among them by spatial disambiguation, the
standard mapping pipeline in deployed systems and in dataset construction
\citep{reddy2025docannot}. \textbf{Output probability} is the untrained confidence baseline, the
grounder's mean probability over the coordinates it emits, used as one scalar with no calibrator.
\label{sec:setup-metrics}

The confidences are scored two ways, each answering a different question. AUROC and expected
calibration error (ECE) are reported for every configuration in each of the six settings, one
grounder condition on one evaluation dataset; Brier and AUPRC are in Appendix~\ref{app:tables}.
Precision and coverage are how grounder conditions are compared, since they do not depend on a
label whose base rate moves with the grounder. \textbf{They are always reported as a pair at the
same threshold}, coverage the fraction auto-accepted and precision the accuracy of what was
accepted, the risk--coverage trade-off of selective prediction
\citep{srinivasan2024selective}.
\section{Results}
\label{sec:results}

\subsection{Localizing the evidence}
\label{sec:res-grounder}

We first state the accuracy of the object being scored. Fine-tuning raises localization recall from
$0.421$ to $0.852$ in distribution, from $0.019$ to $0.586$ on FormNLU, and from $0.421$ to
$0.699$ on VRDU; it beats OCR string mapping in two of three settings by reducing wrong regions
rather than merely increasing found ones (Appendix~\ref{app:grounder}). The frozen model is almost
unusable on FormNLU, localizing about one field in fifty. Since the extractor is fixed, value
correctness is unchanged; what changes below is the grounder and the calibration signal it exposes.

\subsection{The two confidences across settings}
\label{sec:res-system}

\begin{table}[htbp]
\centering\footnotesize
\caption{Main results. Each cell is AUROC\,/\,ECE, mean over 5 seeds. Configurations are defined in \S\ref{sec:method-calibrator}; M4 is the all-feature setting, and \emph{Output prob.} is the untrained floor of \S\ref{sec:setup-baselines}, which has no value-confidence entry because the extractor's log-probabilities were not stored. Evidence-confidence figures are read down a column and not across backbones, because the label is defined by the grounder under test.}
\label{tab:main}
\setlength{\tabcolsep}{4pt}
\begin{tabular}{lcccccc}
\toprule
& \multicolumn{3}{c}{frozen} & \multicolumn{3}{c}{fine-tuned} \\
\cmidrule(lr){2-4}\cmidrule(l){5-7}
& ID & FormNLU & VRDU & ID & FormNLU & VRDU \\
\midrule
\multicolumn{7}{l}{\emph{value confidence}} \\
\midrule
Output prob. & --- & --- & --- & --- & --- & --- \\
M1 & .730/.008 & .451/.075 & .690/.295 & .793/.012 & .531/.155 & .699/.446 \\
M2 & .773/.010 & .510/.044 & .687/.290 & .906/.017 & .614/.111 & .769/.341 \\
M3 & .914/.049 & .543/.220 & .795/.512 & .924/.036 & .591/.195 & .814/.360 \\
\textbf{M4} & .914/.039 & .622/.308 & .793/.481 & .940/.031 & .611/.193 & .783/.354 \\
\midrule
\multicolumn{7}{l}{\emph{evidence confidence}} \\
\midrule
Output prob. & .409/.367 & .416/.849 & .513/.212 & .704/.070 & .399/.206 & .510/.164 \\
M1 & .681/.012 & .618/.321 & .545/.131 & .759/.008 & .644/.181 & .714/.160 \\
M2 & .848/.007 & .974/.037 & .856/.124 & .797/.014 & .778/.170 & .765/.142 \\
M3 & .906/.057 & .925/.010 & .772/.214 & .861/.029 & .598/.236 & .697/.236 \\
\textbf{M4} & .918/.053 & .980/.006 & .859/.174 & .867/.026 & .636/.210 & .739/.229 \\
\bottomrule
\end{tabular}
\end{table}

Table~\ref{tab:main} reports both confidences in all six settings, from the untrained output-probability
floor to M4. The floor is the grounder's mean coordinate-token probability, used without a calibrator;
for evidence it falls below random ordering in three settings, so confidence has to be learned rather
than read off generation probability. In distribution, M4 reaches $0.914$ and $0.940$ AUROC for value on the frozen and fine-tuned
grounders, with ECE at most $0.053$ across the four cells. The evidence figures, $0.918$ and
$0.867$, are not comparable across grounders: fine-tuning lifts that base rate from $0.421$ to
$0.852$, so the wrong locations that remain are fewer and harder to spot which reflects in slight reduction in AUROC.

One OOD cell is a base-rate trap rather than a real perfomance win. Frozen FormNLU evidence has only a $0.019$ positive rate, so M4 can
score almost everything low and obtain AUROC $0.980$ and ECE $0.006$ while auto-accepting only
$1\%$ of fields. AUPRC has the same issue because its floor is the base rate, and stricter IoU
labels lower that base rate without changing the system (Appendix~\ref{app:iou}). We therefore also report
coverage with precision throughout which is a better perfomance indicator for actual deployment, more details in \S\ref{sec:res-finetune}. Feature groups also show the usual tension: they improve ranking
while sometimes worsening calibration; because scalar recalibration can repair ECE but not AUROC, we
select configurations by ranking and recalibrate afterwards (Appendix~\ref{app:temp}).

\subsection{What the confidences are built from}
\label{sec:res-built}

Internals and external agreement help different targets. On the frozen in-distribution backbone,
internals lift value confidence from $0.730$ to $0.914$ and evidence from $0.681$ to $0.906$; external
agreement lifts evidence by $0.167$ but value by only $0.043$ (Table~\ref{tab:main}). This matches its role: it is a post-localization check run with OCR over the grounder proposed location, not the localization mechanism. The two heads are not proxies. The prompt gives the claimed value with the page, so the pass is a
joint function of value and image. Even among the frozen grounder's $8{,}854$ mislocalized fields,
value confidence reaches $0.918$ AUROC, almost the same as on correctly localized fields; internals
supply the signal, while external agreement supplies none (Appendix~\ref{app:decoupling}). Conversely,
external agreement helps value more after fine-tuning ($0.113$ rather than $0.043$), because checking
whether the value lies in the proposed box only becomes meaningful once boxes are usually right. Value confidence therefore does not need the location to be right, which invites verification of extracted values instead of
locating: a verify-only model reaches similar value confidence but returns no evidence to show a
reviewer and no second target to calibrate (Appendix~\ref{app:verify}).

\subsection{Where those features stop working}
\label{sec:res-features}
\begin{figure}[t]
\centering
\includegraphics[width=0.98\linewidth]{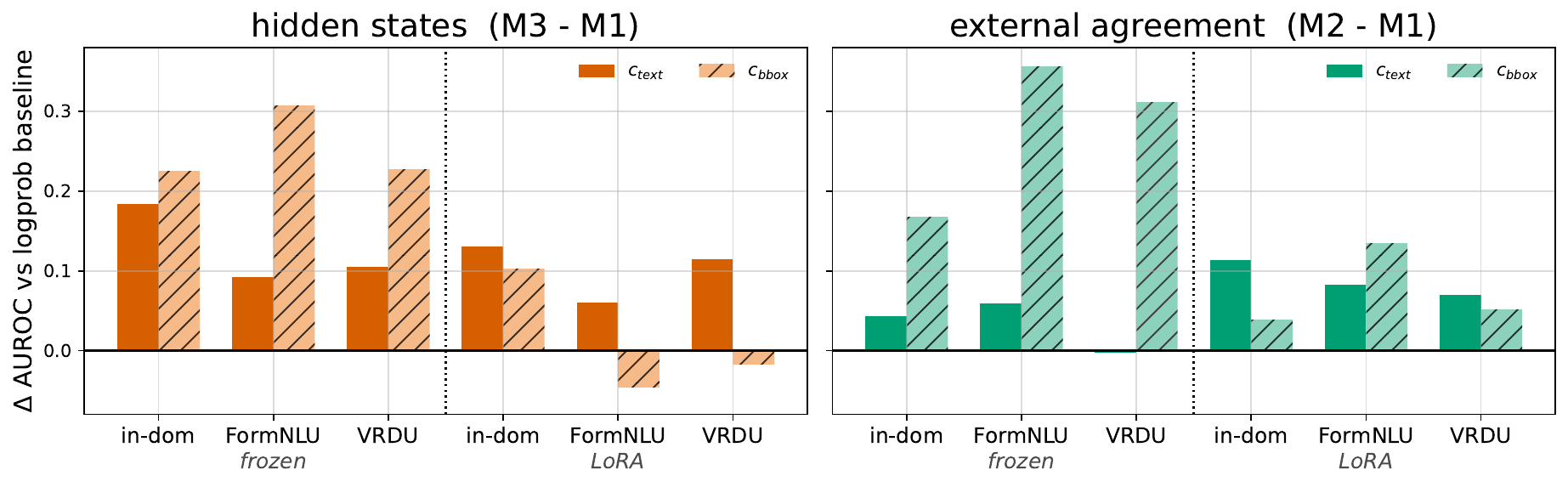}
\caption{Feature contributions over M1. Internals dominate in distribution but turn negative for
fine-tuned OOD evidence confidence; external agreement is weak in distribution and strongest under
shift.}
\label{fig:features}
\end{figure}

The calibrator can fail to transfer even when the grounder localizes well. Figure~\ref{fig:features} plots $\text{M3}-\text{M1}$ for internals and $\text{M2}-\text{M1}$ for
external agreement. Under a frozen backbone, internals survive the shift; after fine-tuning, the same
difference turns negative on both OOD datasets, while external agreement becomes the dominant signal.
Thus the best fine-tuned OOD evidence configuration uses no internals: M2 reaches $0.778$ and
$0.765$ against M4's $0.636$ and $0.739$. RoI features add little by themselves; dropping them moves
AUROC by at most $0.010$ in distribution and inconsistently OOD, while dropping hidden states costs
more in eleven of twelve cells (Appendix~\ref{app:tables}).

The reversal is not a metric artefact: Brier improves from the untrained floor to M4 in distribution,
but OOD improves to M2 and reverses when internals enter (Appendix~\ref{app:tables}). Nor are the
features destroyed. A linear model fitted on target labels recovers fine-tuned evidence at $0.911$
AUROC, while the in-distribution-fitted model reads $0.493$; the answer remains in the representation,
but the learned direction has moved (Appendix~\ref{app:probes}). A verify-only prompt shows the same
OOD value-confidence loss, so the failure is transfer of the confidence model, not localization alone
(Appendix~\ref{app:verify}). Finally, value confidence is also lower for fields sharing an attribute
type with a sibling in the schema, by $0.050$--$0.133$ AUROC in all six settings
(Appendix~\ref{app:twin}).

\subsection{What fine-tuning fixes, and what it does not}
\label{sec:res-finetune}

\begin{figure}[t]
\centering
\includegraphics[width=0.98\linewidth]{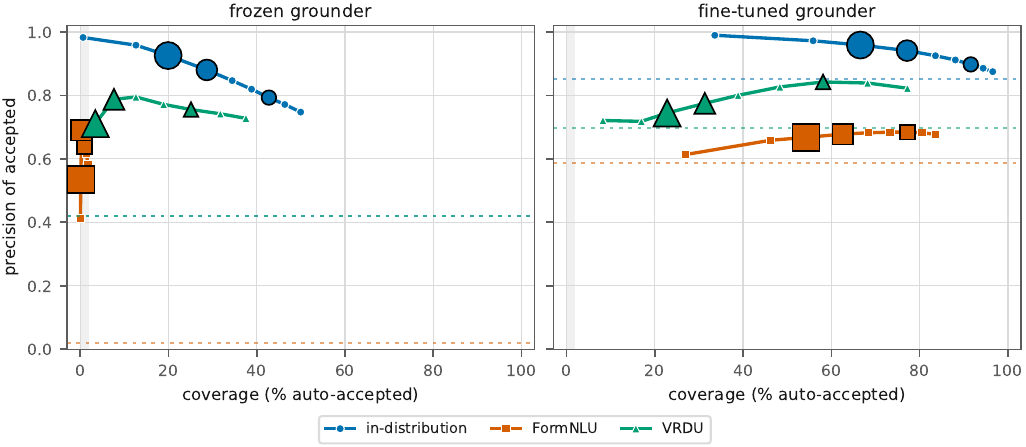}
\caption{Evidence precision--coverage sweeps for frozen and fine-tuned grounders. Each point is a
confidence threshold, running from $0.30$ at the right to $0.99$ at the left; rings mark $0.50$,
$0.80$ and $0.90$, larger for stricter. Dashed lines are accept-everything base rates; the shaded
band marks coverage below $2\%$, where precision is computed over too few decisions to read.}
\label{fig:threshold}
\end{figure}

Fine-tuning creates an operating range, as Figure~\ref{fig:threshold} shows. At threshold $0.80$,
the frozen grounder accepts $29\%$ of fields in distribution, $8\%$ on VRDU and none on FormNLU; at
$0.50$, the highest threshold at which it still accepts enough to judge, it accepts $43\%$, $25\%$
and $1\%$. The fine-tuned grounder accepts $92\%$, $58\%$ and $77\%$ at $0.50$ with better precision
in every case, and still accepts $77\%$ in distribution at threshold $0.80$ with $0.941$ precision
(Table~\ref{tab:operating}). The frozen system therefore offers almost no operating choice, the same
coverage-collapse failure seen in selective VQA \citep{srinivasan2024selective}.

Value confidence gains less, adding $26$ coverage points on FormNLU and $16$ on VRDU at
$0.02$--$0.03$ lower precision, because the extractor being scored is unchanged.

Fine-tuning does not remove the need for target calibration. Without relabelling, evidence
confidence reads $0.641$ on FormNLU and $0.675$ on VRDU, while continuing the in-distribution
calibrator on about $120$ labelled target fields raises them to $0.905$ and $0.829$, around $80\%$
of the full-label gain (Table~\ref{tab:adapt}). Fitting from scratch is worse at every budget, so
the in-distribution calibrator remains a useful starting point even when it transfers poorly.
\subsection{Ablations: Changing the extractor, and removing the VLM}
\label{sec:res-ablations}
A plug-in framework should survive changes to the parts it plugs into. We replace the
extractor, then remove the VLM-localizer entirely, and hold everything else fixed.

\begin{table}[htbp]
\centering\footnotesize
\caption{A different extractor. \emph{base} is the base rate of the target in that row. It
changes with the extractor, so each row is read against its own base rate. \emph{P\,/\,cov} is precision and coverage at confidence $0.90$, coverage as a
percentage of annotated fields.}
\label{tab:extractor}
\setlength{\tabcolsep}{4pt}
\begin{tabular}{lccccccccc}
\toprule
& \multicolumn{3}{c}{ID} & \multicolumn{3}{c}{FormNLU} & \multicolumn{3}{c}{VRDU} \\
\cmidrule(lr){2-4}\cmidrule(lr){5-7}\cmidrule(l){8-10}
Extractor & base & AUROC/ECE & P\,/\,cov & base & AUROC/ECE & P\,/\,cov & base & AUROC/ECE & P\,/\,cov \\
\midrule
\multicolumn{10}{l}{\emph{value confidence}} \\
Gemini 3 Flash   & .600 & .940/.031 & .945/45 & .772 & .611/.193 & .819/34 & .864 & .783/.354 & .977/16 \\
Claude Haiku 4.5 & .426 & .956/.044 & .904/35 & .540 & .692/.156 & .725/28 & .607 & .753/.125 & .898/16 \\
\midrule
\multicolumn{10}{l}{\emph{evidence confidence}} \\
Gemini 3 Flash   & .852 & .867/.026 & .959/67 & .586 & .636/.210 & .668/54 & .699 & .739/.229 & .745/23 \\
Claude Haiku 4.5 & .832 & .861/.026 & .957/62 & .554 & .646/.211 & .636/47 & .625 & .716/.166 & .729/23 \\
\bottomrule
\end{tabular}
\end{table}

\paragraph{A different extractor.}
Table~\ref{tab:extractor} reports the same calibrator over two extractors: the reference Gemini 3
Flash, and Claude Haiku 4.5 \citep{anthropic2025claudehaiku45}, which extracts less accurately. The localizer, the configuration, the
splits and the seeds are identical in both rows. The calibrator behaves the same way under both.
In distribution it discriminates and calibrates well whichever extractor produced the values; out
of distribution both degrade, on the same sets and in the same direction. The axis along which the
calibrator fails is the document domain, not the extractor, and a weaker extractor does not move
that boundary. Within this pattern the two heads separate. Evidence confidence is almost
unaffected by the change of extractor. Value confidence follows the accuracy of the extractor it
scores, and the cost appears as coverage rather than as ranking: under the weaker extractor the
value head ranks no worse, but at a fixed precision target it accepts a smaller share of fields.
Ranking metrics alone would therefore invite the wrong conclusion, and we report the operating
point beside them.

\begin{table}[htbp]
\centering\footnotesize
\caption{No VLM. Each cell is precision\,/\,coverage at that confidence threshold, coverage as a
percentage of annotated fields.}
\label{tab:ocr}
\setlength{\tabcolsep}{4pt}
\begin{tabular}{lcccccc}
\toprule
& \multicolumn{2}{c}{ID} & \multicolumn{2}{c}{FormNLU} & \multicolumn{2}{c}{VRDU} \\
\cmidrule(lr){2-3}\cmidrule(lr){4-5}\cmidrule(l){6-7}
Threshold & OCR & M4 & OCR & M4 & OCR & M4 \\
\midrule
\multicolumn{7}{l}{\emph{evidence confidence}} \\
0.50 & .810/61 & .898/92 & .789/76 & .684/77 & .738/82 & .842/58 \\
0.90 & .934/21 & .959/67 & .828/35 & .668/54 & .902/9 & .745/23 \\
\midrule
\multicolumn{7}{l}{\emph{value confidence}} \\
0.50 & .822/68 & .884/62 & .846/87 & .817/74 & .936/81 & .954/49 \\
0.90 & .924/13 & .945/45 & .808/27 & .819/34 & .807/3 & .977/16 \\
\bottomrule
\end{tabular}
\end{table}

\paragraph{No VLM.}
The feature analysis of \S\ref{sec:res-built} and \S\ref{sec:res-features} shows that an external
OCR signal adds to the confidence, and that it adds most where the internal features weaken. That
invites the opposite question: what if there is no VLM-localizer at all? We evaluate the arm that answers it.
Evidence is located by the OCR value matching of \S\ref{sec:setup-baselines}, and a calibrator over
nine scalars derived from the region it returns produces both confidences. Four describe the
match, namely its fuzzy agreement with the extracted value, how many times the value occurs on the
page, and whether it was found as an exact run of words or only approximately; three describe the
region, namely its area, its aspect ratio and the number of words inside it; two describe the
value, namely its length in tokens and in characters. Nothing in this pipeline generates a box or
reads a hidden state. The question it settles is how close such a system comes to one that reads a
VLM from the inside.

In distribution the fine-tuned grounder with M4 is ahead on both targets: it is more precise at
both thresholds, and for evidence confidence it also accepts more fields. Out of distribution the
result is mixed. On FormNLU the OCR baseline is more precise for evidence confidence at both
thresholds, and for value confidence at $0.50$ it is both more precise and accepts more fields. On
VRDU neither arm leads: the OCR baseline accepts more at the lower threshold, and M4 accepts more at the
higher one. FormNLU is a printed form whose values appear on the page as written, which is the case
value matching handles best, and this is where the baseline stays competitive.

\section{Deployment}
\label{sec:deployment}

VERDICT is a framework with a choice of feature configuration, not a single fixed model, and which configuration to
run depends on constraints a practitioner has: whether the grounder can be fine-tuned,
whether an OCR pass is available, whether any target labels can be produced.
Table~\ref{tab:deploy} maps each of those situations to the configuration we would run and what it
delivers, every row traceable to a result above rather than inferred. These are field-level results, while a deployment decides per document: the natural
use is to accept a document only when the fields it depends on clear both thresholds, and to route
the rest to review with their evidence highlighted
\citep{geifman2017selective,srinivasan2024selective}. We do not evaluate that aggregation here. Two rules follow. Thresholds do not transfer, so a new dataset needs the same labelled target fields
used for warm-starting; and precision is only meaningful with coverage beside it. Where compute is constrained, Qwen3-VL-2B is a reasonable alternative to try: the same
configurations run in both conditions and the frozen-to-fine-tuned pattern reproduces
(Appendix~\ref{app:secondbackbone}).

\begin{table}[htbp]
\centering\footnotesize
\caption{Deployment guidance; every row is traceable to a result above.}
\label{tab:deploy}
\setlength{\tabcolsep}{3pt}
\begin{tabular}{p{0.24\linewidth}p{0.27\linewidth}p{0.39\linewidth}}
\toprule
Situation & Configuration & What to expect \\
\midrule
In distribution & Fine-tuned grounder, M4 & Evidence $0.898$ precision over $92\%$ coverage; value $0.884$ over $62\%$ (\S\ref{sec:res-finetune}) \\
OOD, no target labels & Log-probs + external agreement, no internals & $0.778$/$0.765$ AUROC vs. $0.636$/$0.739$ for M4 (\S\ref{sec:res-features}) \\
OOD, ${\sim}120$ target fields labelled & Warm-start ID calibrator & Evidence confidence $0.641\!\to\!0.905$ and $0.675\!\to\!0.829$ (\S\ref{sec:res-finetune}) \\
Grounder cannot be tuned & Frozen backbone, OCR features & No usable range: $43\%$, $1\%$, $25\%$ coverage at threshold $0.50$ and less above (\S\ref{sec:res-finetune}) \\
No OCR pass & Model internals only & Adequate ID; evidence confidence degrades sharply OOD (\S\ref{sec:res-features}) \\
\bottomrule
\end{tabular}
\end{table}

\section{Conclusion}
\label{sec:conclusion}


VERDICT shows that outputs from a fixed black-box document extractor can be assessed without
modifying the extractor itself. Given the document image and the extractor's JSON output, VERDICT
uses a small open-weight VLM for evidence localization and calibrates value/evidence confidence to
estimate which fields can be trusted. Fine-tuning the VLM improves the evidence available to the
calibrator, which expands VERDICT's selective-prediction operating range for both value and evidence
confidence across ID and OOD datasets, while the frozen setting often loses usable
coverage at stricter thresholds. The results also show that feature choice matters: internal VLM
features are strongest in distribution, but their advantage does not reliably transfer under shift;
optional external agreement and roughly $120$ labelled target fields can recover much of the lost confidence. These findings suggest that evidence localization, feature design, and
lightweight target calibration must be evaluated together when deciding which extractor outputs can
be accepted and which should be routed for review. Our study is limited to one localizer model family: VERDICT consumes only the image plus JSON, and the calibrator behaves the same under a second, weaker extractor, but other VLM based localizer architectures
remain untested. Field existence is read from the extractor's abstention rather than
predicted, and value fabrication is too rare here to study.

\bibliographystyle{plainnat}
\bibliography{refs}


\newpage
\appendix
\section{The grounder}
\label{app:grounder}

\begin{table}[htbp]
\centering\small
\caption{The grounder, before any confidence is attached. \emph{OCR mapping} is the no-model
baseline of \S\ref{sec:setup-baselines}, scored on the same fields at the same IoU; it does not
depend on the backbone, so one row serves both grounder conditions. Metrics are as defined in \S\ref{sec:setup-metrics}; a field left unlocalized costs recall and not
precision. Bold marks the best system per column.}
\label{tab:grounder}
\begin{tabular}{lrrrrrr}
\toprule
& \multicolumn{2}{c}{ID} & \multicolumn{2}{c}{FormNLU} & \multicolumn{2}{c}{VRDU} \\
\cmidrule(lr){2-3}\cmidrule(lr){4-5}\cmidrule(l){6-7}
& P & R & P & R & P & R \\
\midrule
\multicolumn{7}{l}{\emph{fields correctly localized}} \\
OCR mapping           & 0.611 & 0.565 & \textbf{0.687} & \textbf{0.686} & 0.665 & 0.665 \\
Grounder, frozen      & 0.425 & 0.421 & 0.019 & 0.019 & 0.379 & 0.421 \\
Grounder, fine-tuned  & \textbf{0.853} & \textbf{0.852} & 0.651 & 0.586 & \textbf{0.740} & \textbf{0.699} \\
\midrule
\multicolumn{7}{l}{\emph{end-to-end: value and evidence both correct}} \\
Grounder, frozen      & \multicolumn{2}{c}{0.445} & \multicolumn{2}{c}{0.032} & \multicolumn{2}{c}{0.493} \\
Grounder, fine-tuned  & \multicolumn{2}{c}{\textbf{0.714}} & \multicolumn{2}{c}{\textbf{0.678}} & \multicolumn{2}{c}{\textbf{0.757}} \\
\bottomrule
\end{tabular}
\end{table}

A grounding model is only worth using if it beats the OCR-mapping baseline. That baseline is strong,
and beats the \emph{frozen} grounder everywhere, by $14$ points in distribution and $66$ on FormNLU.
Fine-tuning reverses this in two settings of three, leading by $29$ points in distribution and
matching the baseline on VRDU, $0.699$ recall against $0.665$, at clearly higher precision, $0.740$
against $0.665$. It loses on FormNLU, $0.586$ against $0.686$, a fixed-format printed form whose
values appear verbatim, which is the best case for string search. Where a text layer exists and
values are copied verbatim, OCR mapping is a genuine competitor; where the extractor normalises the
value it fails structurally rather than at the margin: in distribution it finds no match at all for
$7.5\%$ of fields, and for $688$ of the $1{,}263$ fields of one currency column. This is why the OCR
feature is strongest out of distribution in \S\ref{sec:res-features}: it imports the judgement of an
independent localizer that is competitive there, exactly where the grounder's own has degraded.

\section{IoU threshold robustness}
\label{app:iou}

The location label uses IoU $\geq 0.4$. Re-deriving it at $0.5$ and retraining the reported
configuration with five seeds moves the in-distribution primary metrics by at most $0.010$ AUROC and
$0.006$ ECE (frozen $0.918 \to 0.913$, ECE $0.053 \to 0.050$; LoRA $0.867 \to 0.858$, ECE
$0.026 \to 0.031$) --- within or barely above seed spread. No in-distribution conclusion depends on the
choice. As noted in \S\ref{sec:res-finetune}, AUPRC flags apparent shifts in four of six settings
here purely because tightening the criterion lowers the base rate.

\section{Prior-art comparison and shared-trunk neutrality}
\label{app:priorwork}

No published system localizes evidence and calibrates confidence over it. Table~\ref{tab:priorwork}
sets out which of the properties in \S\ref{sec:intro} each of the closest systems has. Work on
localization fills the first column and leaves the rest empty; work on confidence fills the second
and leaves the first empty.

\begin{table}[htbp]
\centering\footnotesize
\caption{Structural comparison with the closest prior systems. A dash means the property is absent,
not that the system is worse: these are different tasks. The document visual grounding row stands for
that line of work as a whole rather than a single paper. ExtractConf does construct regions, from OCR
token matching, but only as a means of computing features, and it never scores whether they are
right; neither ApplyBoard nor BICR returns a region at all.}
\label{tab:priorwork}
\begin{tabular}{@{}l >{\raggedright\arraybackslash}p{2.1cm} c
                     >{\raggedright\arraybackslash}p{2.0cm}
                     >{\raggedright\arraybackslash}p{3.1cm}@{}}
\toprule
System & evidence localized & calibrated & targets scored & features the calibrator reads \\
\midrule
LMDX \citep{perot2024lmdx}                    & all fields, one pass  & ---          & ---                & --- \\
Document visual grounding \citep{pm2026perception} & one field per query & ---     & ---                & --- \\
ExtractConf \citep{kumar2026extractconf}      & ---                   & $\checkmark$ & value              & API log-probabilities, OCR, image quality \\
ApplyBoard \citep{macdonald2025embedding}      & ---                   & $\checkmark$ & value              & model internals, text only \\
BICR \citep{khanmohammadi2026bicr}            & ---                   & $\checkmark$ & answer             & model internals \\
\midrule
VERDICT                                       & all fields, one pass  & $\checkmark$ & value and evidence & internals, OCR, log-probabilities \\
\bottomrule
\end{tabular}
\end{table}

Splitting one score into two is not itself new: detection separates classification confidence from
localization quality by regressing box overlap \citep{jiang2018iounet}. Two properties are
ours alone. No prior system scores a second target about the region it produced,
because no confidence system produces a region of its own. And no prior system reads a grounding
model's internals, because the two literatures do not meet: those that ground do not calibrate, and
those that calibrate work on a model that was asked for an answer rather than for a location.

Training each head on an independent trunk rather than a shared one changes AUROC by less than the
seed spread in every setting, so the shared trunk is a parameter-efficiency choice that does not
affect any result.

\section{Second backbone}
\label{app:secondbackbone}

A reduced four-configuration grid was re-run on Qwen3-VL-2B in both frozen and LoRA conditions, five
seeds each (40 runs). In distribution the smaller backbone tracks the larger closely on both heads
(reported M4 configuration, value confidence $0.931$ against $0.940$; location $0.868$ against
$0.867$), and out of distribution it is competitive-to-better on VRDU (value confidence $0.844$ against $0.783$) and
within seed spread of it on FormNLU ($0.566$ against $0.611$). Location confidence is not compared
across backbones: $y_{\mathrm{bbox}}$ is defined by the grounder under test, and the two backbones
localize at different rates on the same pages ($0.343$ against $0.586$ on FormNLU), so the two
columns describe different classification problems.
We draw no scale conclusion from this: the two backbones differ in the number of layers spanned by
the mid/late hidden-state pair as well as in size, so the comparison is not controlled.
Tables~\ref{tab:2b-text} and~\ref{tab:2b-bbox} give the full grid.

\begin{table}[htbp]
\centering\footnotesize
\caption{Second backbone (Qwen3-VL-2B), value confidence (\ctext{}). Four configurations, both arms, mean\,$\pm$\,std over 5 seeds. Read as replication of the \emph{within-backbone} contrasts; the two backbones differ in depth as well as size, so this is not a controlled scale comparison.}
\label{tab:2b-text}
\setlength{\tabcolsep}{4pt}
\begin{tabular}{lcccc}
\toprule
Config & AUROC & ECE & AUPRC & Brier \\
\midrule
\multicolumn{5}{l}{\emph{frozen, In-distribution}, base rate 0.5995} \\
\midrule
M1 & 0.7183\,$\pm$\,0.0012 & 0.0081\,$\pm$\,0.0022 & 0.7497\,$\pm$\,0.0010 & 0.1838\,$\pm$\,0.0001 \\
M2 & 0.7440\,$\pm$\,0.0005 & 0.0099\,$\pm$\,0.0032 & 0.7678\,$\pm$\,0.0011 & 0.1786\,$\pm$\,0.0003 \\
M3 & 0.8973\,$\pm$\,0.0022 & 0.0386\,$\pm$\,0.0144 & 0.9170\,$\pm$\,0.0026 & 0.1237\,$\pm$\,0.0021 \\
\textbf{M4} & 0.8989\,$\pm$\,0.0014 & 0.0414\,$\pm$\,0.0196 & 0.9192\,$\pm$\,0.0017 & 0.1227\,$\pm$\,0.0022 \\
\midrule
\multicolumn{5}{l}{\emph{frozen, FormNLU}, base rate 0.7720} \\
\midrule
M1 & 0.4785\,$\pm$\,0.0035 & 0.0441\,$\pm$\,0.0031 & 0.7670\,$\pm$\,0.0029 & 0.1839\,$\pm$\,0.0009 \\
M2 & 0.5085\,$\pm$\,0.0038 & 0.0282\,$\pm$\,0.0086 & 0.7750\,$\pm$\,0.0047 & 0.1771\,$\pm$\,0.0007 \\
M3 & 0.5250\,$\pm$\,0.0237 & 0.1277\,$\pm$\,0.0268 & 0.7730\,$\pm$\,0.0120 & 0.1964\,$\pm$\,0.0107 \\
\textbf{M4} & 0.5982\,$\pm$\,0.0223 & 0.1281\,$\pm$\,0.0394 & 0.8153\,$\pm$\,0.0111 & 0.1930\,$\pm$\,0.0156 \\
\midrule
\multicolumn{5}{l}{\emph{frozen, VRDU}, base rate 0.8643} \\
\midrule
M1 & 0.7058\,$\pm$\,0.0034 & 0.2477\,$\pm$\,0.0032 & 0.9359\,$\pm$\,0.0014 & 0.1597\,$\pm$\,0.0021 \\
M2 & 0.7191\,$\pm$\,0.0044 & 0.2582\,$\pm$\,0.0136 & 0.9328\,$\pm$\,0.0031 & 0.1691\,$\pm$\,0.0074 \\
M3 & 0.8060\,$\pm$\,0.0051 & 0.4651\,$\pm$\,0.0374 & 0.9606\,$\pm$\,0.0017 & 0.3584\,$\pm$\,0.0369 \\
\textbf{M4} & 0.8162\,$\pm$\,0.0096 & 0.4220\,$\pm$\,0.0531 & 0.9628\,$\pm$\,0.0021 & 0.3152\,$\pm$\,0.0488 \\
\midrule
\multicolumn{5}{l}{\emph{LoRA, In-distribution}, base rate 0.5995} \\
\midrule
M1 & 0.7789\,$\pm$\,0.0011 & 0.0183\,$\pm$\,0.0087 & 0.8046\,$\pm$\,0.0018 & 0.1724\,$\pm$\,0.0007 \\
M2 & 0.8927\,$\pm$\,0.0010 & 0.0148\,$\pm$\,0.0075 & 0.8960\,$\pm$\,0.0019 & 0.1163\,$\pm$\,0.0010 \\
M3 & 0.9165\,$\pm$\,0.0020 & 0.0380\,$\pm$\,0.0085 & 0.9288\,$\pm$\,0.0034 & 0.1106\,$\pm$\,0.0016 \\
\textbf{M4} & 0.9306\,$\pm$\,0.0018 & 0.0407\,$\pm$\,0.0097 & 0.9414\,$\pm$\,0.0020 & 0.0995\,$\pm$\,0.0020 \\
\midrule
\multicolumn{5}{l}{\emph{LoRA, FormNLU}, base rate 0.7720} \\
\midrule
M1 & 0.5812\,$\pm$\,0.0094 & 0.1172\,$\pm$\,0.0137 & 0.8254\,$\pm$\,0.0034 & 0.1933\,$\pm$\,0.0028 \\
M2 & 0.5848\,$\pm$\,0.0022 & 0.1342\,$\pm$\,0.0112 & 0.8217\,$\pm$\,0.0011 & 0.1964\,$\pm$\,0.0045 \\
M3 & 0.5365\,$\pm$\,0.0109 & 0.1480\,$\pm$\,0.0336 & 0.7709\,$\pm$\,0.0044 & 0.1973\,$\pm$\,0.0117 \\
\textbf{M4} & 0.5660\,$\pm$\,0.0207 & 0.1201\,$\pm$\,0.0104 & 0.7883\,$\pm$\,0.0054 & 0.1861\,$\pm$\,0.0051 \\
\midrule
\multicolumn{5}{l}{\emph{LoRA, VRDU}, base rate 0.8643} \\
\midrule
M1 & 0.7375\,$\pm$\,0.0034 & 0.3845\,$\pm$\,0.0121 & 0.9334\,$\pm$\,0.0028 & 0.2490\,$\pm$\,0.0102 \\
M2 & 0.7730\,$\pm$\,0.0040 & 0.3119\,$\pm$\,0.0174 & 0.9529\,$\pm$\,0.0018 & 0.2007\,$\pm$\,0.0146 \\
M3 & 0.8185\,$\pm$\,0.0051 & 0.3055\,$\pm$\,0.0411 & 0.9589\,$\pm$\,0.0010 & 0.2153\,$\pm$\,0.0399 \\
\textbf{M4} & 0.8442\,$\pm$\,0.0087 & 0.2693\,$\pm$\,0.0538 & 0.9666\,$\pm$\,0.0028 & 0.1873\,$\pm$\,0.0387 \\
\bottomrule
\end{tabular}
\end{table}

\begin{table}[htbp]
\centering\footnotesize
\caption{Second backbone (Qwen3-VL-2B), location confidence (\cbbox{}). Four configurations, both arms, mean\,$\pm$\,std over 5 seeds. Read as replication of the \emph{within-backbone} contrasts; the two backbones differ in depth as well as size, so this is not a controlled scale comparison.}
\label{tab:2b-bbox}
\setlength{\tabcolsep}{4pt}
\begin{tabular}{lcccc}
\toprule
Config & AUROC & ECE & AUPRC & Brier \\
\midrule
\multicolumn{5}{l}{\emph{frozen, In-distribution}, base rate 0.2069} \\
\midrule
M1 & 0.7263\,$\pm$\,0.0007 & 0.0135\,$\pm$\,0.0012 & 0.3916\,$\pm$\,0.0008 & 0.1468\,$\pm$\,0.0001 \\
M2 & 0.8925\,$\pm$\,0.0004 & 0.0100\,$\pm$\,0.0007 & 0.6238\,$\pm$\,0.0019 & 0.1051\,$\pm$\,0.0002 \\
M3 & 0.9201\,$\pm$\,0.0019 & 0.0279\,$\pm$\,0.0053 & 0.7535\,$\pm$\,0.0059 & 0.0909\,$\pm$\,0.0013 \\
\textbf{M4} & 0.9458\,$\pm$\,0.0019 & 0.0288\,$\pm$\,0.0057 & 0.8119\,$\pm$\,0.0043 & 0.0775\,$\pm$\,0.0016 \\
\midrule
\multicolumn{5}{l}{\emph{frozen, FormNLU}, base rate 0.0040 --- \emph{25 positives; carries no claim}} \\
\midrule
M1 & 0.6066\,$\pm$\,0.0046 & 0.1075\,$\pm$\,0.0006 & 0.0068\,$\pm$\,0.0003 & 0.0208\,$\pm$\,0.0002 \\
M2 & 0.9770\,$\pm$\,0.0002 & 0.0248\,$\pm$\,0.0002 & 0.1194\,$\pm$\,0.0091 & 0.0117\,$\pm$\,0.0003 \\
M3 & 0.8516\,$\pm$\,0.0216 & 0.0020\,$\pm$\,0.0015 & 0.0688\,$\pm$\,0.0193 & 0.0040\,$\pm$\,0.0001 \\
\textbf{M4} & 0.9813\,$\pm$\,0.0040 & 0.0034\,$\pm$\,0.0016 & 0.1727\,$\pm$\,0.0286 & 0.0042\,$\pm$\,0.0004 \\
\midrule
\multicolumn{5}{l}{\emph{frozen, VRDU}, base rate 0.2056} \\
\midrule
M1 & 0.5119\,$\pm$\,0.0059 & 0.0749\,$\pm$\,0.0065 & 0.2087\,$\pm$\,0.0015 & 0.1701\,$\pm$\,0.0004 \\
M2 & 0.8274\,$\pm$\,0.0022 & 0.0639\,$\pm$\,0.0032 & 0.4260\,$\pm$\,0.0030 & 0.1317\,$\pm$\,0.0006 \\
M3 & 0.7615\,$\pm$\,0.0066 & 0.0987\,$\pm$\,0.0212 & 0.4791\,$\pm$\,0.0191 & 0.1497\,$\pm$\,0.0050 \\
\textbf{M4} & 0.8255\,$\pm$\,0.0212 & 0.0905\,$\pm$\,0.0189 & 0.5744\,$\pm$\,0.0477 & 0.1347\,$\pm$\,0.0060 \\
\midrule
\multicolumn{5}{l}{\emph{LoRA, In-distribution}, base rate 0.8074} \\
\midrule
M1 & 0.7660\,$\pm$\,0.0023 & 0.0148\,$\pm$\,0.0142 & 0.9207\,$\pm$\,0.0013 & 0.1287\,$\pm$\,0.0009 \\
M2 & 0.7927\,$\pm$\,0.0035 & 0.0212\,$\pm$\,0.0201 & 0.9305\,$\pm$\,0.0012 & 0.1215\,$\pm$\,0.0027 \\
M3 & 0.8578\,$\pm$\,0.0038 & 0.0306\,$\pm$\,0.0122 & 0.9603\,$\pm$\,0.0012 & 0.1104\,$\pm$\,0.0014 \\
\textbf{M4} & 0.8680\,$\pm$\,0.0054 & 0.0318\,$\pm$\,0.0146 & 0.9626\,$\pm$\,0.0016 & 0.1055\,$\pm$\,0.0014 \\
\midrule
\multicolumn{5}{l}{\emph{LoRA, FormNLU}, base rate 0.3429} \\
\midrule
M1 & 0.7198\,$\pm$\,0.0011 & 0.2257\,$\pm$\,0.0151 & 0.4812\,$\pm$\,0.0016 & 0.2664\,$\pm$\,0.0081 \\
M2 & 0.8242\,$\pm$\,0.0008 & 0.1883\,$\pm$\,0.0197 & 0.6020\,$\pm$\,0.0015 & 0.1982\,$\pm$\,0.0091 \\
M3 & 0.8040\,$\pm$\,0.0130 & 0.1561\,$\pm$\,0.0483 & 0.6631\,$\pm$\,0.0194 & 0.2070\,$\pm$\,0.0259 \\
\textbf{M4} & 0.8726\,$\pm$\,0.0176 & 0.1429\,$\pm$\,0.0396 & 0.7239\,$\pm$\,0.0415 & 0.1679\,$\pm$\,0.0210 \\
\midrule
\multicolumn{5}{l}{\emph{LoRA, VRDU}, base rate 0.5729} \\
\midrule
M1 & 0.6939\,$\pm$\,0.0008 & 0.1712\,$\pm$\,0.0146 & 0.6973\,$\pm$\,0.0010 & 0.2389\,$\pm$\,0.0068 \\
M2 & 0.7624\,$\pm$\,0.0017 & 0.1404\,$\pm$\,0.0145 & 0.7363\,$\pm$\,0.0020 & 0.1821\,$\pm$\,0.0087 \\
M3 & 0.7460\,$\pm$\,0.0195 & 0.1321\,$\pm$\,0.0224 & 0.7256\,$\pm$\,0.0168 & 0.2090\,$\pm$\,0.0090 \\
\textbf{M4} & 0.7690\,$\pm$\,0.0115 & 0.1350\,$\pm$\,0.0098 & 0.7260\,$\pm$\,0.0107 & 0.1785\,$\pm$\,0.0079 \\
\bottomrule
\end{tabular}
\end{table}

\section{Where the internal representation moves}
\label{app:probes}

This appendix supports the claim in \S\ref{sec:res-features} that fine-tuning moves where the
location signal sits rather than removing it.

These measurements do \emph{not} refit M4. They fit a linear model, on the same features the
calibrator reads, reduced to $64$ components. A linear model is used deliberately: it makes ``where
the signal sits'' a set of weights that can be compared between two fits, which a multi-layer
calibrator would not. The conclusions are therefore about the features, not about the calibrator
architecture. Both arms are compared on the \emph{same} features, so the datasets and the number of
features are held fixed.

Two fits are compared by the cosine of their weights, which is $1$ when they agree, $0$ when they
are unrelated and negative when they oppose; over $64$ weights, two unrelated fits score about
$0.100$ in magnitude. Five seeds, with the two classes balanced by subsampling, over annotated
fields only, which is the population every other number in this paper is computed on.
Features are standardised and reduced, both fitted on the calibrator's training split alone,
because weights over $2560$ features estimated from FormNLU's $117$ correct boxes would be noise.

Two consequences follow. Weights can only be compared this way for a linear model, so
Table~\ref{tab:rotation} describes what is linearly readable from the features, not what a
multi-layer calibrator does with them. Table~\ref{tab:oracle} carries no such restriction, and
refitting M4 itself on FormNLU reaches $0.967$ against the $0.911$ a linear model reads from
one feature group, so the linear measurement understates what the features hold rather than
overstating it.

\begin{table}[htbp]
\centering\footnotesize
\caption{Agreement between the $y_{\mathrm{bbox}}$ weights fitted in distribution and the weights fitted
out of distribution. Lower means they moved further apart, and $0.100$ is what two unrelated fits score.
Fine-tuning moves them further in all six internal cells, while the two controls stay where they
were.}
\label{tab:rotation}
\begin{tabular}{lcccc}
\toprule
& \multicolumn{2}{c}{FormNLU} & \multicolumn{2}{c}{VRDU} \\
\cmidrule(lr){2-3}\cmidrule(l){4-5}
Feature & frozen & LoRA & frozen & LoRA \\
\midrule
hidden, mid layer  & $+0.226$ & $-0.060$ & $+0.261$ & $+0.201$ \\
hidden, late layer & $+0.162$ & $-0.099$ & $+0.218$ & $-0.017$ \\
visual RoI         & $+0.344$ & $+0.176$ & $+0.318$ & $-0.081$ \\
\midrule
log-probabilities \emph{(control)} & $+0.307$ & $+0.307$ & $+0.583$ & $+0.867$ \\
OCR \emph{(control)}               & $+0.967$ & $+0.897$ & $+0.996$ & $+0.908$ \\
\bottomrule
\end{tabular}
\end{table}

\begin{table}[htbp]
\centering\footnotesize
\caption{$y_{\mathrm{bbox}}$ AUROC on FormNLU, from weights fitted in distribution against weights fitted
on FormNLU itself. The second column of each pair is five-fold: the weights are fitted on four
folds of FormNLU and scored on the fifth, so no record is scored by weights fitted on it. It is
still not available in deployment, since it needs the labels the confidence is meant to predict,
but it shows the information is present in the features. The gap is much larger under LoRA on
every internal feature, and near zero for both controls.}
\label{tab:oracle}
\begin{tabular}{lcccc}
\toprule
& \multicolumn{2}{c}{frozen} & \multicolumn{2}{c}{LoRA} \\
\cmidrule(lr){2-3}\cmidrule(l){4-5}
Feature & in-distribution fit & FormNLU fit & in-distribution fit & FormNLU fit \\
\midrule
hidden, mid layer  & $0.846$ & $0.917$ & $0.493$ & $0.911$ \\
hidden, late layer & $0.766$ & $0.929$ & $0.544$ & $0.892$ \\
visual RoI         & $0.914$ & $0.944$ & $0.395$ & $0.872$ \\
\midrule
log-probabilities \emph{(control)} & $0.556$ & $0.542$ & $0.634$ & $0.650$ \\
OCR \emph{(control)}               & $0.971$ & $0.970$ & $0.797$ & $0.799$ \\
\bottomrule
\end{tabular}
\end{table}

Repeating the first measurement with shuffled labels gives $0.420$ to $0.526$ across the four
settings, so the splits do not leak. These measurements show the weights move and the signal
survives. They do not establish that the movement \emph{causes} the calibration loss.

\section{Value confidence when the box is wrong}
\label{app:decoupling}

Records are split on the outcome: those the grounder mislocated ($y_{\mathrm{bbox}}=0$) and those it
located correctly. Value confidence is then scored inside each subset separately, over annotated
fields, 5 seeds. Signal surviving in the first subset cannot have come from a successful
localization, because there was none.

\begin{table}[htbp]
\centering\footnotesize
\caption{Value confidence on the fields the grounder mislocated. AUROC on \ctext{}, mean over 5 seeds, computed only over records with $y_{\mathrm{bbox}}=0$. \emph{M1} is the log-probability floor, \emph{M3} adds hidden states and no OCR, so $\Delta$ is what the internal representation contributes on records whose box was wrong. The last column is M3 on the complementary subset. Value confidence does not track localisation success.}
\label{tab:decoupling}
\setlength{\tabcolsep}{4pt}
\begin{tabular}{lrcccc}
\toprule
Setting & \multicolumn{1}{c}{box wrong} & M1 & M3 & $\Delta$ & M3, box right \\
\midrule
frozen, in-distribution & 8,854 (57.9\%) & 0.762 & 0.918 & $+0.157$ & 0.908 \\
frozen, FormNLU & 6,112 (98.1\%) & 0.449 & 0.540 & $+0.091$ & 0.740 \\
frozen, VRDU & 290 (57.9\%) & 0.695 & 0.788 & $+0.093$ & 0.813 \\
LoRA, in-distribution & 2,256 (14.8\%) & 0.686 & 0.773 & $+0.087$ & 0.942 \\
LoRA, FormNLU & 2,579 (41.4\%) & 0.533 & 0.543 & $+0.010$ & 0.710 \\
LoRA, VRDU & 151 (30.1\%) & 0.762 & 0.804 & $+0.042$ & 0.820 \\
\bottomrule
\end{tabular}
\end{table}

\texttt{match\_score} is near-constant across the mislocated subset, since the value is not inside
the predicted box, so adding it should change nothing there. It does not: M4 is within $0.003$ of
M3 on both frozen in distribution and frozen VRDU, and below it under LoRA. None of the gain in
Table~\ref{tab:decoupling} is OCR.

Two limits. The frozen rows are the stronger evidence, because under LoRA a wrong box and a wrong
value are correlated: the \ctext{} base rate on the mislocated subset falls to $0.295$ from $0.600$
overall, against $0.515$ from $0.600$ under frozen. And LoRA/FormNLU is the one row where hidden
states add nothing, on the dataset where value confidence is near chance for every subset anyway
(Appendix~\ref{app:twin}). Subset AUROCs should be compared with each other and not with a figure
computed over all records, which draws on cross-subset pairs that neither subset contains.

\section{Judging instead of locating}
\label{app:verify}

The same frozen model is asked to answer Y or N for each extracted value rather than to locate it,
one pass per document, with features read at the answer token. The calibrator is fitted on the same
in distribution documents and scored on the same records, using the same two architectures and five
seeds. There is no box, so this arm has no location confidence and no OCR feature.

\begin{table}[htbp]
\centering\footnotesize
\caption{Judging instead of locating. Value confidence \ctext{} AUROC, frozen backbone, mean over 5 seeds, on the same records as Table~\ref{tab:main}. \emph{untrained} is the model's own answer used directly; \emph{scalars} and \emph{hidden} are the same two calibrators fitted on each arm's features. Verification is ahead in distribution ($0.929$ against $0.910$) and behind on VRDU; on FormNLU both arms are at chance, so that column compares two models neither of which is usable. Verification yields no box, so it has no location confidence and no OCR feature at all.}
\label{tab:verify}
\setlength{\tabcolsep}{4pt}
\begin{tabular}{llccc}
\toprule
 & & In-distribution & FormNLU & VRDU \\
\midrule
verification & untrained & 0.729 & 0.655 & 0.607 \\
 & scalars & 0.743 & 0.655 & 0.610 \\
 & hidden & 0.929 & 0.501 & 0.799 \\
grounding & scalars & 0.730 & 0.451 & 0.690 \\
 & hidden & 0.910 & 0.513 & 0.823 \\
 & M4 & 0.914 & 0.622 & 0.793 \\
\bottomrule
\end{tabular}
\end{table}

The model's own Y/N answer is correct on $69.1\%$ of fields against a $60.0\%$ base rate, so what
Table~\ref{tab:verify} measures is what its internal states carry, not the judgment it states.

\section{Schema duplication, by setting}
\label{app:twin}
This diagnostic checks whether fields become harder when the schema contains multiple fields of
the same broad type, such as several dates or several names. Such ``twin'' fields can create
schema-level ambiguity: the extracted value may look plausible, but correspond to the wrong field.

\begin{figure}[h]
\centering
\includegraphics[width=\linewidth]{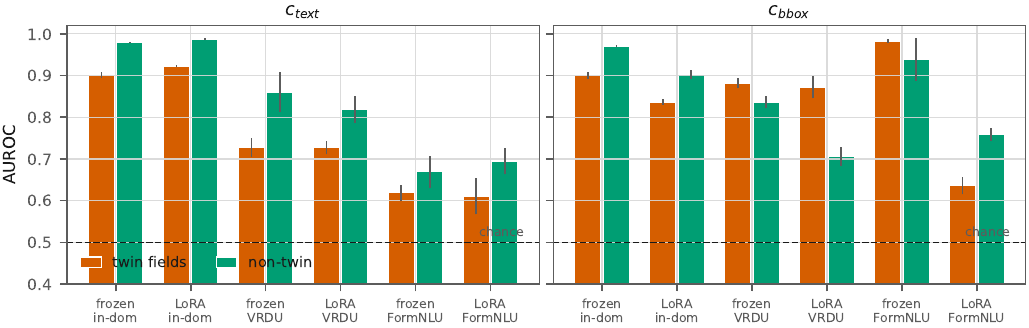}
\caption{Fields sharing a core attribute type with a sibling in the same schema (``twin'': two dates,
three names) against fields with no sibling, for the reported configuration. Value confidence is worse
on twin fields in all six settings; location confidence shows no consistent effect, which is what
scopes the diagnostic to \ctext{}. Labelling was fixed before results were inspected. The
frozen/FormNLU location cell is omitted --- at a base rate of $0.010$ the ranking metric is not a
measurement.}
\label{fig:twin}
\end{figure}

As shown in Figure~\ref{fig:twin}, twin fields mainly affect value confidence, not evidence
confidence. Across the reported settings, value confidence is consistently worse on twin fields than
on non-twin fields, suggesting that the risk comes from schema-level ambiguity rather than visual
evidence localisation. Evidence confidence does not show a consistent twin penalty, so the
diagnostic is most relevant for interpreting \ctext{} and for deciding which fields should be routed
to review. In practice, schemas with repeated field types should be treated as higher-risk extraction
settings, even when the evidence box appears plausible.

\section{Repairing calibration with a single scalar}
\label{app:temp}

A single temperature fitted on $100$ labelled target fields improves ECE in all eight
out-of-distribution settings, by $0.004$ to $0.159$. The repair is partial: residual ECE is $0.069$
to $0.354$, against $0.026$ to $0.053$ in distribution.

Because the transformation is monotone it cannot reorder fields, so AUROC and coverage at a fixed
threshold are unchanged. We verified this rather than assuming it: $399$ of $400$ cells are exactly
invariant. This is what licenses selecting a configuration on ranking and repairing calibration
afterwards (\S\ref{sec:res-system}).

It is a convenience rather than a capability gain. Anyone able to label $100$ target fields could
instead re-tune the decision threshold on those same fields and obtain the same operating point
without refitting anything.

\section{Operating points}
\label{app:operating}

\begin{table}[htbp]
\centering\footnotesize
\caption{Operating points. Each cell is precision\,/\,coverage at that confidence threshold,
coverage as a percentage of annotated fields. The frozen columns are the point of the table:
raising the threshold empties the accepted set, leaving a precision computed over almost no
decisions, so the frozen grounder offers no operating range a deployment could choose from.}
\label{tab:operating}
\setlength{\tabcolsep}{4pt}
\begin{tabular}{lcccccc}
\toprule
& \multicolumn{2}{c}{ID} & \multicolumn{2}{c}{FormNLU} & \multicolumn{2}{c}{VRDU} \\
\cmidrule(lr){2-3}\cmidrule(lr){4-5}\cmidrule(l){6-7}
Threshold & frozen & tuned & frozen & tuned & frozen & tuned \\
\midrule
\multicolumn{7}{l}{\emph{evidence confidence}} \\
0.50 & .793/43 & .898/92 & .638/1 & .684/77 & .755/25 & .842/58 \\
0.70 & .847/34 & .925/84 & .686/1 & .682/68 & .795/13 & .800/39 \\
0.80 & .880/29 & .941/77 & .690/0 & .678/63 & .788/8 & .775/31 \\
0.90 & .926/20 & .959/67 & .536/0 & .668/54 & .711/3 & .745/23 \\
\midrule
\multicolumn{7}{l}{\emph{value confidence}} \\
0.50 & .854/63 & .884/62 & .836/48 & .817/74 & .986/33 & .954/49 \\
0.70 & .887/55 & .912/56 & .860/25 & .832/55 & .991/22 & .965/30 \\
0.80 & .908/48 & .927/52 & .855/14 & .835/45 & .991/16 & .970/23 \\
0.90 & .933/37 & .945/45 & .777/5 & .819/34 & .998/10 & .977/16 \\
\bottomrule
\end{tabular}
\end{table}

\begin{figure}[htbp]
\centering
\includegraphics[width=\linewidth]{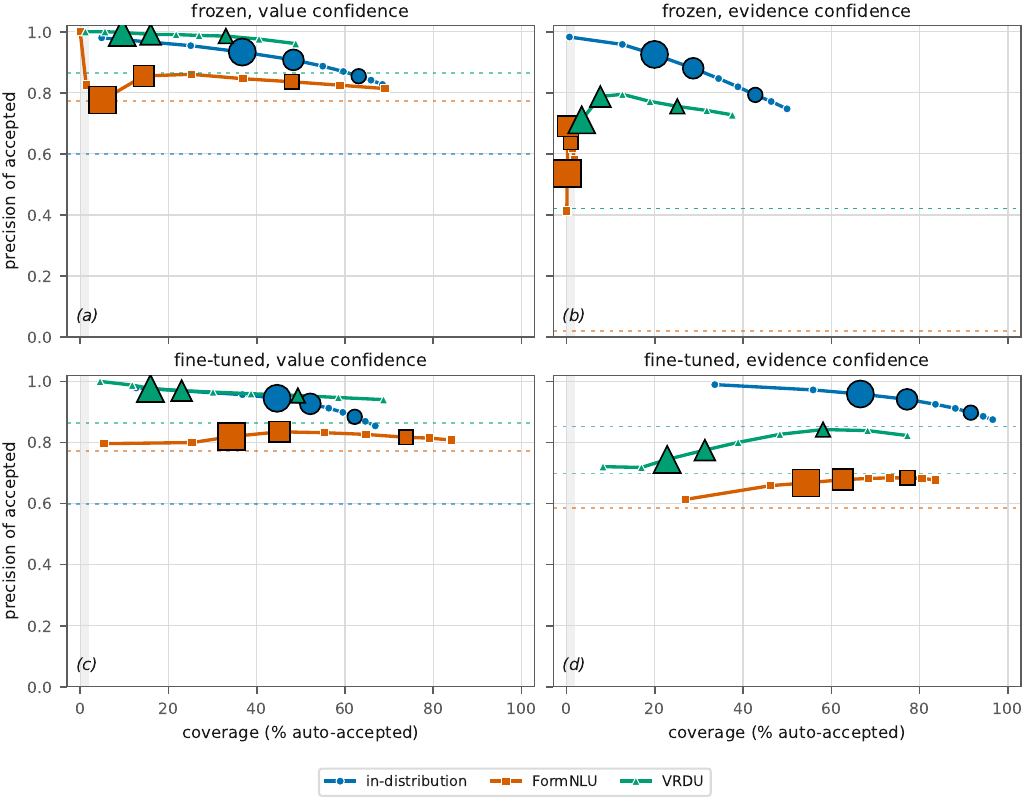}
\caption{The threshold sweep of Figure~\ref{fig:threshold} over both targets and both grounder
conditions, on the same axes throughout. Reading the dashed base-rate lines down a column shows the
label asymmetry of \S\ref{sec:method-task} directly: value confidence has the same base rates in
both rows, because the extractor is fixed, while evidence confidence does not, because that label is
defined by the grounder under test.}
\label{fig:operating-full}
\end{figure}

Table~\ref{tab:operating} gives precision and coverage at four thresholds for both grounder
conditions. It supports the claim in \S\ref{sec:res-finetune} that the frozen grounder has no
operating range: its coverage on evidence confidence falls to $29\%$, $0\%$ and $8\%$ at a threshold
of $0.80$, so the precision reported beside those figures describes almost no decisions. The
fine-tuned grounder still accepts $77\%$, $63\%$ and $31\%$ at the same threshold.

\section{Adaptation cost}
\label{app:adapt}

\begin{figure}[htbp]
\centering
\includegraphics[width=\linewidth]{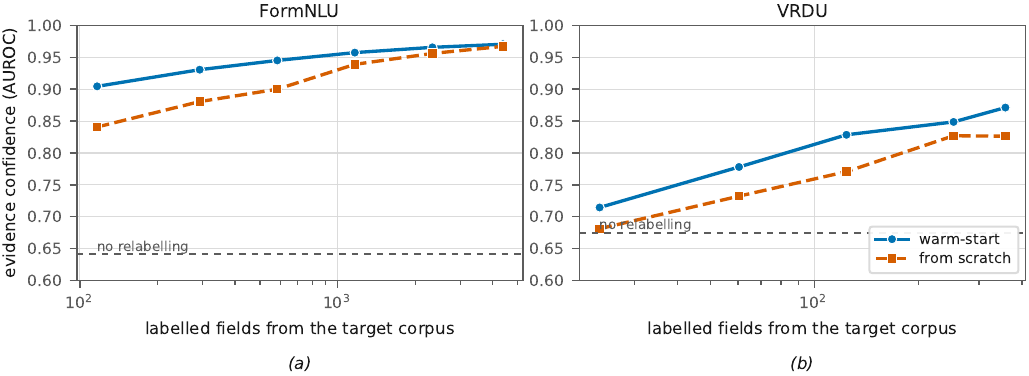}
\caption{Evidence-confidence AUROC against the number of labelled fields drawn from the target
dataset. \emph{Warm-start} continues from the in-distribution calibrator; \emph{from scratch} fits on
the target labels alone. The dashed line is the in-distribution calibrator applied to the target
dataset with no relabelling. The two panels use different horizontal ranges because the datasets
differ in size, so positions should not be compared across them.}
\label{fig:adapt}
\end{figure}
\begin{table}[htbp]
\centering\footnotesize
\caption{Adaptation cost. Evidence- and value-confidence AUROC against the number of labelled
fields drawn from the target dataset, mean over 5 seeds. \emph{scratch} fits on the target labels
alone, \emph{warm} continues from the in-distribution calibrator, and the $0$ row is that
calibrator applied with no relabelling. Warm-start is ahead of scratch at every budget on both
targets. The budgets quoted in \S\ref{sec:res-finetune} are the $117$-field row on FormNLU and the
$124$-field row on VRDU. These are the numbers behind Figure~\ref{fig:adapt}.}
\label{tab:adapt}
\setlength{\tabcolsep}{6pt}
\begin{tabular}{rr cc cc}
\toprule
& & \multicolumn{2}{c}{evidence confidence} & \multicolumn{2}{c}{value confidence} \\
\cmidrule(lr){3-4}\cmidrule(l){5-6}
docs & fields & scratch & warm & scratch & warm \\
\midrule
\multicolumn{6}{l}{\emph{FormNLU}} \\
\midrule
0    & 0     & ---   & 0.641          & ---   & 0.605          \\
10   & 117   & 0.841 & \textbf{0.905} & 0.743 & \textbf{0.799} \\
25   & 292   & 0.881 & 0.931          & 0.836 & 0.872          \\
50   & 584   & 0.900 & 0.945          & 0.881 & 0.908          \\
100  & 1166  & 0.939 & 0.958          & 0.913 & 0.935          \\
200  & 2332  & 0.956 & 0.966          & 0.942 & 0.949          \\
375  & 4376  & 0.967 & 0.971          & 0.951 & 0.960          \\
\midrule
\multicolumn{6}{l}{\emph{VRDU}} \\
\midrule
0    & 0     & ---   & 0.675          & ---   & 0.841          \\
10   & 24    & 0.681 & 0.714          & 0.555 & 0.732          \\
25   & 60    & 0.732 & 0.778          & 0.646 & 0.766          \\
50   & 124   & 0.771 & \textbf{0.829} & 0.695 & 0.745          \\
100  & 253   & 0.827 & 0.849          & 0.727 & 0.749          \\
140  & 358   & 0.826 & 0.871          & 0.736 & 0.769          \\
\bottomrule
\end{tabular}
\end{table}
Figure~\ref{fig:adapt} plots evidence-confidence AUROC against the labelling budget. The calibrator is fitted on labelled fields drawn from the target dataset and scored on a held-out
set of documents from the same dataset, with nested draws so that a larger budget always contains a
smaller one. Every budget is trained for a fixed number of gradient steps rather than a fixed number
of epochs, since fixed epochs would give the largest budget far more optimisation than the smallest
and confound budget with training length. No early stopping is used and no validation split is
carved from the target, so nothing is selected on the set being reported.

This measures the cost of refitting the calibrator, not out-of-distribution performance. Once the
calibrator has seen labelled fields from the target dataset, that dataset is in distribution for it.
What remains out of distribution is the grounder, which is never retrained here.

\section{Full result tables}
\label{app:tables}
All ten configurations in all six settings, both targets, mean\,$\pm$\,std over five seeds. These
are the tables behind every contrast in \S\ref{sec:results}; each cell traces to one entry of the
seed aggregate, and no number here is recomputed for presentation. Each setting is headed by its base rate, because AUPRC is bounded below by it and two settings
with different base rates are not comparable without them.

\begin{table}[htbp]
\centering\footnotesize
\caption{Configuration definitions for the full grids. M2 is the strongest scalar/external policy and M4 is the all-feature VERDICT instantiation used in the main tables. The suffixes name one feature that is present or absent relative to that family. Here ``full'' means hidden states plus visual RoI features.}
\label{tab:config-defs}
\setlength{\tabcolsep}{4pt}
\begin{tabular}{lccccc}
\toprule
Variant & Arch & Scalars & \texttt{match\_score} & \texttt{ambiguity\_count} & Feature set \\
\midrule
M1 & MLP\textsubscript{A} & 6 & --- & --- & --- \\
M2-match & MLP\textsubscript{A} & 7 & $\checkmark$ & --- & --- \\
M2-ambig & MLP\textsubscript{A} & 7 & --- & $\checkmark$ & --- \\
M2 & MLP\textsubscript{A} & 8 & $\checkmark$ & $\checkmark$ & --- \\
M3 & MLP\textsubscript{B} & 4 & --- & --- & full \\
M4-match & MLP\textsubscript{B} & 5 & $\checkmark$ & --- & full \\
M4-ambig & MLP\textsubscript{B} & 5 & --- & $\checkmark$ & full \\
M4 & MLP\textsubscript{B} & 6 & $\checkmark$ & $\checkmark$ & full \\
M4-hidden & MLP\textsubscript{B} & 4 & --- & --- & hidden \\
M4-roi & MLP\textsubscript{B} & 4 & --- & --- & roi \\
\bottomrule
\end{tabular}
\end{table}

\begin{table}[htbp]
\centering\footnotesize
\caption{Full results, frozen backbone, value confidence (\ctext{}). Mean\,$\pm$\,std over 5 seeds. AUPRC is bounded below by the base rate, which is given for each setting; only lift over that floor is comparable between settings.}
\label{tab:full-frozen-text}
\setlength{\tabcolsep}{4pt}
\begin{tabular}{lcccc}
\toprule
Config & AUROC & ECE & AUPRC & Brier \\
\midrule
\multicolumn{5}{l}{\emph{In-distribution}, base rate 0.5995} \\
\midrule
M1 & 0.7301\,$\pm$\,0.0003 & 0.0079\,$\pm$\,0.0010 & 0.7582\,$\pm$\,0.0005 & 0.1820\,$\pm$\,0.0001 \\
M2-match & 0.7496\,$\pm$\,0.0005 & 0.0080\,$\pm$\,0.0011 & 0.7809\,$\pm$\,0.0005 & 0.1783\,$\pm$\,0.0001 \\
M2-ambig & 0.7584\,$\pm$\,0.0005 & 0.0059\,$\pm$\,0.0003 & 0.7860\,$\pm$\,0.0005 & 0.1766\,$\pm$\,0.0002 \\
M2 & 0.7730\,$\pm$\,0.0008 & 0.0095\,$\pm$\,0.0014 & 0.8038\,$\pm$\,0.0011 & 0.1729\,$\pm$\,0.0002 \\
M3 & 0.9139\,$\pm$\,0.0033 & 0.0488\,$\pm$\,0.0109 & 0.9273\,$\pm$\,0.0023 & 0.1126\,$\pm$\,0.0035 \\
M4-match & 0.9144\,$\pm$\,0.0043 & 0.0481\,$\pm$\,0.0124 & 0.9284\,$\pm$\,0.0032 & 0.1127\,$\pm$\,0.0032 \\
M4-ambig & 0.9124\,$\pm$\,0.0041 & 0.0413\,$\pm$\,0.0206 & 0.9267\,$\pm$\,0.0031 & 0.1129\,$\pm$\,0.0019 \\
\textbf{M4} & 0.9143\,$\pm$\,0.0054 & 0.0394\,$\pm$\,0.0172 & 0.9284\,$\pm$\,0.0044 & 0.1115\,$\pm$\,0.0029 \\
M4-hidden & 0.9100\,$\pm$\,0.0057 & 0.0451\,$\pm$\,0.0145 & 0.9235\,$\pm$\,0.0062 & 0.1142\,$\pm$\,0.0032 \\
M4-roi & 0.8917\,$\pm$\,0.0028 & 0.0315\,$\pm$\,0.0070 & 0.9078\,$\pm$\,0.0029 & 0.1223\,$\pm$\,0.0013 \\
\midrule
\multicolumn{5}{l}{\emph{FormNLU}, base rate 0.7720} \\
\midrule
M1 & 0.4505\,$\pm$\,0.0036 & 0.0750\,$\pm$\,0.0090 & 0.7525\,$\pm$\,0.0017 & 0.1916\,$\pm$\,0.0003 \\
M2-match & 0.4478\,$\pm$\,0.0073 & 0.0728\,$\pm$\,0.0069 & 0.7515\,$\pm$\,0.0038 & 0.1840\,$\pm$\,0.0005 \\
M2-ambig & 0.5172\,$\pm$\,0.0011 & 0.0713\,$\pm$\,0.0035 & 0.8063\,$\pm$\,0.0006 & 0.1865\,$\pm$\,0.0011 \\
M2 & 0.5096\,$\pm$\,0.0067 & 0.0441\,$\pm$\,0.0026 & 0.7966\,$\pm$\,0.0046 & 0.1813\,$\pm$\,0.0007 \\
M3 & 0.5428\,$\pm$\,0.0356 & 0.2197\,$\pm$\,0.0843 & 0.7869\,$\pm$\,0.0256 & 0.2473\,$\pm$\,0.0495 \\
M4-match & 0.5524\,$\pm$\,0.0293 & 0.2795\,$\pm$\,0.1080 & 0.7885\,$\pm$\,0.0156 & 0.2896\,$\pm$\,0.0708 \\
M4-ambig & 0.5877\,$\pm$\,0.0317 & 0.2331\,$\pm$\,0.0795 & 0.8149\,$\pm$\,0.0236 & 0.2564\,$\pm$\,0.0479 \\
\textbf{M4} & 0.6218\,$\pm$\,0.0192 & 0.3077\,$\pm$\,0.1044 & 0.8316\,$\pm$\,0.0118 & 0.3034\,$\pm$\,0.0713 \\
M4-hidden & 0.5132\,$\pm$\,0.0328 & 0.1699\,$\pm$\,0.0653 & 0.7578\,$\pm$\,0.0200 & 0.2115\,$\pm$\,0.0293 \\
M4-roi & 0.5024\,$\pm$\,0.0249 & 0.2613\,$\pm$\,0.0778 & 0.7706\,$\pm$\,0.0140 & 0.2973\,$\pm$\,0.0513 \\
\midrule
\multicolumn{5}{l}{\emph{VRDU}, base rate 0.8643} \\
\midrule
M1 & 0.6900\,$\pm$\,0.0029 & 0.2954\,$\pm$\,0.0026 & 0.9240\,$\pm$\,0.0007 & 0.1886\,$\pm$\,0.0018 \\
M2-match & 0.6659\,$\pm$\,0.0029 & 0.2613\,$\pm$\,0.0030 & 0.9072\,$\pm$\,0.0008 & 0.1765\,$\pm$\,0.0016 \\
M2-ambig & 0.6991\,$\pm$\,0.0118 & 0.3196\,$\pm$\,0.0060 & 0.9270\,$\pm$\,0.0025 & 0.2055\,$\pm$\,0.0048 \\
M2 & 0.6872\,$\pm$\,0.0066 & 0.2895\,$\pm$\,0.0064 & 0.9113\,$\pm$\,0.0025 & 0.1947\,$\pm$\,0.0037 \\
M3 & 0.7953\,$\pm$\,0.0114 & 0.5116\,$\pm$\,0.0427 & 0.9597\,$\pm$\,0.0024 & 0.4131\,$\pm$\,0.0336 \\
M4-match & 0.7754\,$\pm$\,0.0123 & 0.4889\,$\pm$\,0.0579 & 0.9556\,$\pm$\,0.0026 & 0.4003\,$\pm$\,0.0683 \\
M4-ambig & 0.7959\,$\pm$\,0.0180 & 0.4987\,$\pm$\,0.0328 & 0.9599\,$\pm$\,0.0043 & 0.3935\,$\pm$\,0.0319 \\
\textbf{M4} & 0.7928\,$\pm$\,0.0124 & 0.4810\,$\pm$\,0.0504 & 0.9597\,$\pm$\,0.0039 & 0.3810\,$\pm$\,0.0533 \\
M4-hidden & 0.8233\,$\pm$\,0.0173 & 0.4046\,$\pm$\,0.0578 & 0.9650\,$\pm$\,0.0034 & 0.3034\,$\pm$\,0.0594 \\
M4-roi & 0.7753\,$\pm$\,0.0077 & 0.4552\,$\pm$\,0.0498 & 0.9555\,$\pm$\,0.0025 & 0.3395\,$\pm$\,0.0451 \\
\bottomrule
\end{tabular}
\end{table}

\begin{table}[htbp]
\centering\footnotesize
\caption{Full results, frozen backbone, location confidence (\cbbox{}). Mean\,$\pm$\,std over 5 seeds. AUPRC is bounded below by the base rate, which is given for each setting; only lift over that floor is comparable between settings.}
\label{tab:full-frozen-bbox}
\setlength{\tabcolsep}{4pt}
\begin{tabular}{lcccc}
\toprule
Config & AUROC & ECE & AUPRC & Brier \\
\midrule
\multicolumn{5}{l}{\emph{In-distribution}, base rate 0.4210} \\
\midrule
M1 & 0.6806\,$\pm$\,0.0007 & 0.0119\,$\pm$\,0.0020 & 0.5548\,$\pm$\,0.0013 & 0.2155\,$\pm$\,0.0002 \\
M2-match & 0.8252\,$\pm$\,0.0002 & 0.0095\,$\pm$\,0.0010 & 0.6950\,$\pm$\,0.0009 & 0.1605\,$\pm$\,0.0001 \\
M2-ambig & 0.7095\,$\pm$\,0.0001 & 0.0128\,$\pm$\,0.0015 & 0.5846\,$\pm$\,0.0011 & 0.2092\,$\pm$\,0.0001 \\
M2 & 0.8476\,$\pm$\,0.0003 & 0.0072\,$\pm$\,0.0012 & 0.7384\,$\pm$\,0.0018 & 0.1533\,$\pm$\,0.0001 \\
M3 & 0.9058\,$\pm$\,0.0024 & 0.0571\,$\pm$\,0.0144 & 0.8660\,$\pm$\,0.0037 & 0.1295\,$\pm$\,0.0025 \\
M4-match & 0.9184\,$\pm$\,0.0016 & 0.0569\,$\pm$\,0.0108 & 0.8823\,$\pm$\,0.0032 & 0.1202\,$\pm$\,0.0019 \\
M4-ambig & 0.9035\,$\pm$\,0.0032 & 0.0486\,$\pm$\,0.0211 & 0.8636\,$\pm$\,0.0050 & 0.1297\,$\pm$\,0.0021 \\
\textbf{M4} & 0.9179\,$\pm$\,0.0056 & 0.0526\,$\pm$\,0.0088 & 0.8825\,$\pm$\,0.0088 & 0.1204\,$\pm$\,0.0032 \\
M4-hidden & 0.8953\,$\pm$\,0.0051 & 0.0491\,$\pm$\,0.0128 & 0.8529\,$\pm$\,0.0065 & 0.1338\,$\pm$\,0.0022 \\
M4-roi & 0.8795\,$\pm$\,0.0043 & 0.0466\,$\pm$\,0.0128 & 0.8259\,$\pm$\,0.0061 & 0.1431\,$\pm$\,0.0023 \\
\midrule
\multicolumn{5}{l}{\emph{FormNLU}, base rate 0.0188} \\
\midrule
M1 & 0.6183\,$\pm$\,0.0036 & 0.3208\,$\pm$\,0.0021 & 0.0263\,$\pm$\,0.0008 & 0.1409\,$\pm$\,0.0017 \\
M2-match & 0.9733\,$\pm$\,0.0005 & 0.0373\,$\pm$\,0.0008 & 0.2901\,$\pm$\,0.0127 & 0.0235\,$\pm$\,0.0002 \\
M2-ambig & 0.6425\,$\pm$\,0.0030 & 0.3237\,$\pm$\,0.0017 & 0.0311\,$\pm$\,0.0014 & 0.1467\,$\pm$\,0.0015 \\
M2 & 0.9743\,$\pm$\,0.0003 & 0.0369\,$\pm$\,0.0011 & 0.2879\,$\pm$\,0.0035 & 0.0244\,$\pm$\,0.0002 \\
M3 & 0.9250\,$\pm$\,0.0186 & 0.0098\,$\pm$\,0.0033 & 0.3698\,$\pm$\,0.0200 & 0.0145\,$\pm$\,0.0004 \\
M4-match & 0.9741\,$\pm$\,0.0144 & 0.0050\,$\pm$\,0.0015 & 0.5370\,$\pm$\,0.0472 & 0.0118\,$\pm$\,0.0008 \\
M4-ambig & 0.9201\,$\pm$\,0.0222 & 0.0124\,$\pm$\,0.0062 & 0.3504\,$\pm$\,0.0386 & 0.0149\,$\pm$\,0.0009 \\
\textbf{M4} & 0.9799\,$\pm$\,0.0048 & 0.0058\,$\pm$\,0.0019 & 0.5416\,$\pm$\,0.0443 & 0.0119\,$\pm$\,0.0008 \\
M4-hidden & 0.8950\,$\pm$\,0.0096 & 0.0316\,$\pm$\,0.0160 & 0.3043\,$\pm$\,0.0298 & 0.0179\,$\pm$\,0.0022 \\
M4-roi & 0.8663\,$\pm$\,0.0212 & 0.0454\,$\pm$\,0.0218 & 0.2007\,$\pm$\,0.0264 & 0.0264\,$\pm$\,0.0058 \\
\midrule
\multicolumn{5}{l}{\emph{VRDU}, base rate 0.4212} \\
\midrule
M1 & 0.5448\,$\pm$\,0.0083 & 0.1312\,$\pm$\,0.0116 & 0.4344\,$\pm$\,0.0060 & 0.2570\,$\pm$\,0.0034 \\
M2-match & 0.8600\,$\pm$\,0.0010 & 0.1254\,$\pm$\,0.0045 & 0.7252\,$\pm$\,0.0030 & 0.1649\,$\pm$\,0.0010 \\
M2-ambig & 0.5374\,$\pm$\,0.0027 & 0.1496\,$\pm$\,0.0037 & 0.4293\,$\pm$\,0.0018 & 0.2665\,$\pm$\,0.0008 \\
M2 & 0.8556\,$\pm$\,0.0014 & 0.1243\,$\pm$\,0.0043 & 0.7129\,$\pm$\,0.0028 & 0.1664\,$\pm$\,0.0025 \\
M3 & 0.7720\,$\pm$\,0.0140 & 0.2143\,$\pm$\,0.0392 & 0.6523\,$\pm$\,0.0192 & 0.2488\,$\pm$\,0.0198 \\
M4-match & 0.8572\,$\pm$\,0.0148 & 0.1768\,$\pm$\,0.0327 & 0.7392\,$\pm$\,0.0288 & 0.2010\,$\pm$\,0.0197 \\
M4-ambig & 0.7672\,$\pm$\,0.0092 & 0.1757\,$\pm$\,0.0356 & 0.6500\,$\pm$\,0.0169 & 0.2318\,$\pm$\,0.0169 \\
\textbf{M4} & 0.8590\,$\pm$\,0.0100 & 0.1742\,$\pm$\,0.0316 & 0.7383\,$\pm$\,0.0172 & 0.2009\,$\pm$\,0.0263 \\
M4-hidden & 0.7002\,$\pm$\,0.0226 & 0.1087\,$\pm$\,0.0173 & 0.6043\,$\pm$\,0.0086 & 0.2261\,$\pm$\,0.0065 \\
M4-roi & 0.7511\,$\pm$\,0.0092 & 0.2044\,$\pm$\,0.0382 & 0.6259\,$\pm$\,0.0148 & 0.2517\,$\pm$\,0.0187 \\
\bottomrule
\end{tabular}
\end{table}

\begin{table}[htbp]
\centering\footnotesize
\caption{Full results, lora backbone, value confidence (\ctext{}). Mean\,$\pm$\,std over 5 seeds. AUPRC is bounded below by the base rate, which is given for each setting; only lift over that floor is comparable between settings.}
\label{tab:full-lora-text}
\setlength{\tabcolsep}{4pt}
\begin{tabular}{lcccc}
\toprule
Config & AUROC & ECE & AUPRC & Brier \\
\midrule
\multicolumn{5}{l}{\emph{In-distribution}, base rate 0.5995} \\
\midrule
M1 & 0.7925\,$\pm$\,0.0015 & 0.0120\,$\pm$\,0.0022 & 0.8136\,$\pm$\,0.0022 & 0.1673\,$\pm$\,0.0004 \\
M2-match & 0.9034\,$\pm$\,0.0012 & 0.0161\,$\pm$\,0.0043 & 0.9064\,$\pm$\,0.0015 & 0.1073\,$\pm$\,0.0011 \\
M2-ambig & 0.8112\,$\pm$\,0.0021 & 0.0112\,$\pm$\,0.0029 & 0.8339\,$\pm$\,0.0032 & 0.1608\,$\pm$\,0.0007 \\
M2 & 0.9059\,$\pm$\,0.0010 & 0.0170\,$\pm$\,0.0036 & 0.9111\,$\pm$\,0.0014 & 0.1051\,$\pm$\,0.0008 \\
M3 & 0.9243\,$\pm$\,0.0019 & 0.0361\,$\pm$\,0.0086 & 0.9384\,$\pm$\,0.0022 & 0.1058\,$\pm$\,0.0015 \\
M4-match & 0.9398\,$\pm$\,0.0015 & 0.0250\,$\pm$\,0.0093 & 0.9501\,$\pm$\,0.0025 & 0.0911\,$\pm$\,0.0015 \\
M4-ambig & 0.9241\,$\pm$\,0.0028 & 0.0323\,$\pm$\,0.0065 & 0.9385\,$\pm$\,0.0037 & 0.1057\,$\pm$\,0.0014 \\
\textbf{M4} & 0.9396\,$\pm$\,0.0031 & 0.0306\,$\pm$\,0.0078 & 0.9497\,$\pm$\,0.0033 & 0.0910\,$\pm$\,0.0030 \\
M4-hidden & 0.9228\,$\pm$\,0.0034 & 0.0414\,$\pm$\,0.0062 & 0.9378\,$\pm$\,0.0032 & 0.1069\,$\pm$\,0.0026 \\
M4-roi & 0.8914\,$\pm$\,0.0029 & 0.0409\,$\pm$\,0.0059 & 0.9081\,$\pm$\,0.0039 & 0.1243\,$\pm$\,0.0025 \\
\midrule
\multicolumn{5}{l}{\emph{FormNLU}, base rate 0.7720} \\
\midrule
M1 & 0.5311\,$\pm$\,0.0075 & 0.1545\,$\pm$\,0.0048 & 0.7802\,$\pm$\,0.0038 & 0.2051\,$\pm$\,0.0014 \\
M2-match & 0.6081\,$\pm$\,0.0016 & 0.1068\,$\pm$\,0.0106 & 0.8162\,$\pm$\,0.0029 & 0.1930\,$\pm$\,0.0036 \\
M2-ambig & 0.5548\,$\pm$\,0.0073 & 0.1600\,$\pm$\,0.0149 & 0.7966\,$\pm$\,0.0061 & 0.2059\,$\pm$\,0.0049 \\
M2 & 0.6135\,$\pm$\,0.0047 & 0.1105\,$\pm$\,0.0034 & 0.8119\,$\pm$\,0.0082 & 0.1920\,$\pm$\,0.0020 \\
M3 & 0.5909\,$\pm$\,0.0170 & 0.1947\,$\pm$\,0.0239 & 0.7872\,$\pm$\,0.0098 & 0.2192\,$\pm$\,0.0163 \\
M4-match & 0.5979\,$\pm$\,0.0314 & 0.1905\,$\pm$\,0.0198 & 0.7929\,$\pm$\,0.0162 & 0.2207\,$\pm$\,0.0091 \\
M4-ambig & 0.6013\,$\pm$\,0.0324 & 0.1684\,$\pm$\,0.0440 & 0.7943\,$\pm$\,0.0186 & 0.2050\,$\pm$\,0.0195 \\
\textbf{M4} & 0.6109\,$\pm$\,0.0452 & 0.1929\,$\pm$\,0.0307 & 0.8086\,$\pm$\,0.0296 & 0.2245\,$\pm$\,0.0181 \\
M4-hidden & 0.5649\,$\pm$\,0.0274 & 0.1856\,$\pm$\,0.0355 & 0.7807\,$\pm$\,0.0175 & 0.2257\,$\pm$\,0.0239 \\
M4-roi & 0.5480\,$\pm$\,0.0320 & 0.1961\,$\pm$\,0.0416 & 0.7875\,$\pm$\,0.0159 & 0.2369\,$\pm$\,0.0259 \\
\midrule
\multicolumn{5}{l}{\emph{VRDU}, base rate 0.8643} \\
\midrule
M1 & 0.6985\,$\pm$\,0.0040 & 0.4458\,$\pm$\,0.0088 & 0.9192\,$\pm$\,0.0038 & 0.3061\,$\pm$\,0.0072 \\
M2-match & 0.7751\,$\pm$\,0.0059 & 0.3174\,$\pm$\,0.0178 & 0.9489\,$\pm$\,0.0042 & 0.2024\,$\pm$\,0.0121 \\
M2-ambig & 0.7151\,$\pm$\,0.0042 & 0.4767\,$\pm$\,0.0188 & 0.9256\,$\pm$\,0.0040 & 0.3363\,$\pm$\,0.0181 \\
M2 & 0.7690\,$\pm$\,0.0028 & 0.3407\,$\pm$\,0.0144 & 0.9470\,$\pm$\,0.0034 & 0.2185\,$\pm$\,0.0099 \\
M3 & 0.8135\,$\pm$\,0.0211 & 0.3599\,$\pm$\,0.0469 & 0.9589\,$\pm$\,0.0057 & 0.2633\,$\pm$\,0.0428 \\
M4-match & 0.8046\,$\pm$\,0.0175 & 0.3470\,$\pm$\,0.0128 & 0.9587\,$\pm$\,0.0038 & 0.2483\,$\pm$\,0.0121 \\
M4-ambig & 0.8168\,$\pm$\,0.0144 & 0.3449\,$\pm$\,0.0718 & 0.9600\,$\pm$\,0.0032 & 0.2535\,$\pm$\,0.0617 \\
\textbf{M4} & 0.7829\,$\pm$\,0.0189 & 0.3535\,$\pm$\,0.0218 & 0.9527\,$\pm$\,0.0050 & 0.2642\,$\pm$\,0.0244 \\
M4-hidden & 0.8038\,$\pm$\,0.0194 & 0.3261\,$\pm$\,0.0602 & 0.9574\,$\pm$\,0.0047 & 0.2442\,$\pm$\,0.0457 \\
M4-roi & 0.7229\,$\pm$\,0.0203 & 0.5128\,$\pm$\,0.0535 & 0.9412\,$\pm$\,0.0066 & 0.4103\,$\pm$\,0.0617 \\
\bottomrule
\end{tabular}
\end{table}

\begin{table}[htbp]
\centering\footnotesize
\caption{Full results, lora backbone, location confidence (\cbbox{}). Mean\,$\pm$\,std over 5 seeds. AUPRC is bounded below by the base rate, which is given for each setting; only lift over that floor is comparable between settings.}
\label{tab:full-lora-bbox}
\setlength{\tabcolsep}{4pt}
\begin{tabular}{lcccc}
\toprule
Config & AUROC & ECE & AUPRC & Brier \\
\midrule
\multicolumn{5}{l}{\emph{In-distribution}, base rate 0.8525} \\
\midrule
M1 & 0.7587\,$\pm$\,0.0036 & 0.0077\,$\pm$\,0.0043 & 0.9398\,$\pm$\,0.0013 & 0.1071\,$\pm$\,0.0005 \\
M2-match & 0.7749\,$\pm$\,0.0049 & 0.0126\,$\pm$\,0.0081 & 0.9448\,$\pm$\,0.0013 & 0.1043\,$\pm$\,0.0012 \\
M2-ambig & 0.7823\,$\pm$\,0.0033 & 0.0106\,$\pm$\,0.0079 & 0.9472\,$\pm$\,0.0009 & 0.1048\,$\pm$\,0.0010 \\
M2 & 0.7971\,$\pm$\,0.0053 & 0.0143\,$\pm$\,0.0046 & 0.9519\,$\pm$\,0.0012 & 0.1018\,$\pm$\,0.0012 \\
M3 & 0.8615\,$\pm$\,0.0049 & 0.0289\,$\pm$\,0.0102 & 0.9714\,$\pm$\,0.0013 & 0.0935\,$\pm$\,0.0018 \\
M4-match & 0.8649\,$\pm$\,0.0039 & 0.0313\,$\pm$\,0.0202 & 0.9721\,$\pm$\,0.0011 & 0.0919\,$\pm$\,0.0028 \\
M4-ambig & 0.8607\,$\pm$\,0.0061 & 0.0265\,$\pm$\,0.0167 & 0.9713\,$\pm$\,0.0015 & 0.0937\,$\pm$\,0.0024 \\
\textbf{M4} & 0.8672\,$\pm$\,0.0053 & 0.0255\,$\pm$\,0.0139 & 0.9725\,$\pm$\,0.0012 & 0.0905\,$\pm$\,0.0018 \\
M4-hidden & 0.8629\,$\pm$\,0.0045 & 0.0324\,$\pm$\,0.0147 & 0.9719\,$\pm$\,0.0010 & 0.0944\,$\pm$\,0.0009 \\
M4-roi & 0.7815\,$\pm$\,0.0078 & 0.0316\,$\pm$\,0.0120 & 0.9491\,$\pm$\,0.0015 & 0.1066\,$\pm$\,0.0009 \\
\midrule
\multicolumn{5}{l}{\emph{FormNLU}, base rate 0.5860} \\
\midrule
M1 & 0.6437\,$\pm$\,0.0021 & 0.1806\,$\pm$\,0.0065 & 0.6666\,$\pm$\,0.0071 & 0.2442\,$\pm$\,0.0023 \\
M2-match & 0.7321\,$\pm$\,0.0028 & 0.1607\,$\pm$\,0.0130 & 0.7298\,$\pm$\,0.0012 & 0.2062\,$\pm$\,0.0029 \\
M2-ambig & 0.6850\,$\pm$\,0.0027 & 0.1918\,$\pm$\,0.0082 & 0.7050\,$\pm$\,0.0039 & 0.2423\,$\pm$\,0.0031 \\
M2 & 0.7778\,$\pm$\,0.0061 & 0.1695\,$\pm$\,0.0094 & 0.7736\,$\pm$\,0.0036 & 0.2028\,$\pm$\,0.0024 \\
M3 & 0.5979\,$\pm$\,0.0167 & 0.2362\,$\pm$\,0.0139 & 0.6181\,$\pm$\,0.0136 & 0.2723\,$\pm$\,0.0057 \\
M4-match & 0.6265\,$\pm$\,0.0324 & 0.2113\,$\pm$\,0.0191 & 0.6275\,$\pm$\,0.0235 & 0.2491\,$\pm$\,0.0118 \\
M4-ambig & 0.5868\,$\pm$\,0.0155 & 0.2388\,$\pm$\,0.0051 & 0.6129\,$\pm$\,0.0143 & 0.2755\,$\pm$\,0.0037 \\
\textbf{M4} & 0.6359\,$\pm$\,0.0197 & 0.2103\,$\pm$\,0.0136 & 0.6387\,$\pm$\,0.0244 & 0.2529\,$\pm$\,0.0058 \\
M4-hidden & 0.6175\,$\pm$\,0.0362 & 0.2441\,$\pm$\,0.0330 & 0.6450\,$\pm$\,0.0256 & 0.2767\,$\pm$\,0.0203 \\
M4-roi & 0.6262\,$\pm$\,0.0120 & 0.1743\,$\pm$\,0.0426 & 0.6435\,$\pm$\,0.0123 & 0.2486\,$\pm$\,0.0192 \\
\midrule
\multicolumn{5}{l}{\emph{VRDU}, base rate 0.6986} \\
\midrule
M1 & 0.7140\,$\pm$\,0.0012 & 0.1595\,$\pm$\,0.0112 & 0.7969\,$\pm$\,0.0011 & 0.1923\,$\pm$\,0.0047 \\
M2-match & 0.7705\,$\pm$\,0.0087 & 0.1289\,$\pm$\,0.0145 & 0.8277\,$\pm$\,0.0085 & 0.1604\,$\pm$\,0.0052 \\
M2-ambig & 0.7102\,$\pm$\,0.0021 & 0.1616\,$\pm$\,0.0066 & 0.7930\,$\pm$\,0.0022 & 0.1921\,$\pm$\,0.0033 \\
M2 & 0.7649\,$\pm$\,0.0042 & 0.1418\,$\pm$\,0.0160 & 0.8166\,$\pm$\,0.0049 & 0.1530\,$\pm$\,0.0084 \\
M3 & 0.6967\,$\pm$\,0.0122 & 0.2364\,$\pm$\,0.0277 & 0.7818\,$\pm$\,0.0082 & 0.2226\,$\pm$\,0.0132 \\
M4-match & 0.7468\,$\pm$\,0.0045 & 0.2009\,$\pm$\,0.0358 & 0.7980\,$\pm$\,0.0044 & 0.1807\,$\pm$\,0.0184 \\
M4-ambig & 0.6986\,$\pm$\,0.0179 & 0.2245\,$\pm$\,0.0379 & 0.7802\,$\pm$\,0.0173 & 0.2155\,$\pm$\,0.0231 \\
\textbf{M4} & 0.7386\,$\pm$\,0.0227 & 0.2295\,$\pm$\,0.0350 & 0.7975\,$\pm$\,0.0130 & 0.2060\,$\pm$\,0.0216 \\
M4-hidden & 0.7053\,$\pm$\,0.0090 & 0.2218\,$\pm$\,0.0436 & 0.7802\,$\pm$\,0.0111 & 0.2230\,$\pm$\,0.0271 \\
M4-roi & 0.6494\,$\pm$\,0.0246 & 0.2894\,$\pm$\,0.0535 & 0.7520\,$\pm$\,0.0244 & 0.2799\,$\pm$\,0.0454 \\
\bottomrule
\end{tabular}
\end{table}

\section{Reproducibility}
\label{app:repro}

The late layer the hidden states are read from sits at $78\%$ of depth. Neither it nor the middle
layer was tuned, and no result in this paper was selected over layer choice.

Every reported cell is the mean over five seeds with the random state reset before each
configuration, so a cell can be reproduced in isolation and comparisons at a fixed seed see identical
batch orders. The location metrics are computed on the same record population used for training; an
independently written re-implementation reproduced them to zero difference. High-dimensional features
are read from a contiguous store verified bit-exact against the originals on 300 randomly drawn
records per file.

\end{document}